\documentclass[11pt,a4paper]{article}
\usepackage{times,latexsym}
\usepackage{url}
\usepackage[T1]{fontenc}

\usepackage[acceptedWithA]{tacl2021v1}
\usepackage{xspace,mfirstuc,tabulary}

\newif\iftaclinstructions
\taclinstructionsfalse % AUTHORS: do NOT set this to true
\iftaclinstructions
\renewcommand{\confidential}{}
\renewcommand{\anonsubtext}{(No author info supplied here, for consistency with
TACL-submission anonymization requirements)}
\newcommand{\instr}
\fi

\iftaclpubformat % this "if" is set by the choice of options

\else

\fi

\usepackage{graphicx}
\usepackage{amsmath} 
\usepackage{amsfonts}
\usepackage{booktabs}
\usepackage{natbib}
\usepackage{subcaption}
\usepackage{hyperref}
\hypersetup{
    colorlinks=true,
    linkcolor=darkblue,
    filecolor=darkblue,      
    urlcolor=darkblue,
    }
\usepackage{xcolor}
\definecolor{goodtxt}{RGB}{0,100,0}
\definecolor{badtxt}{RGB}{153,0,0}
\definecolor{preftxt}{RGB}{96,96,96}
\newcommand{\pref}[1]{\textcolor{preftxt}{#1}}
\newcommand{\good}[1]{\textcolor{goodtxt}{#1}}
\newcommand{\bad}[1]{\textcolor{badtxt}{#1}}

\title{Early Detection and Reduction of Memorisation for Domain Adaptation and Instruction Tuning}
\author{%
  Dean L.~Slack \quad Noura Al Moubayed\\[4pt]
  Durham University, U.K.\\
  \texttt{\{dean.l.slack,noura.al-moubayed\}@durham.ac.uk}
}
\date{}

\begin{document}
\maketitle

\begin{abstract}
Although large language models excel across many tasks, they can memorise training data and thereby expose private or copyrighted text.  Most defences target the pre-training stage, leaving memorisation during fine-tuning—especially for domain adaptation and instruction tuning—poorly understood.  We fine-tune Pythia, Llama3, and Mistral models spanning 1.4B–70B parameters on common evaluation datasets and track verbatim memorisation throughout training.  We find that memorisation increases dramatically in the first few epochs, often significantly before either validation perplexity or evaluation performance is optimised.  We use a simple but effective \(n\)-gram memorisation score which reliably precedes verbatim memorisation; using it as an early-stopping criterion mitigates memorisation with minimal performance loss.  Further, we introduce an \(n\)-gram–aware loss regulariser and show that it reduces memorisation across all model families tested by up to 40\% while minimising evaluation performance trade-offs when compared to an existing memorisation mitigation strategy. These results yield practical, scalable insights into memorisation dynamics during language model fine-tuning.
\end{abstract}

\section{Introduction}
Large Language Models (LLMs) have become increasingly powerful, achieving remarkable performance across diverse tasks and domains as they scale from millions to trillions of parameters \citep{brown2020language, fedus2022switch}. Transformer-based architectures have propelled significant advancements in Natural Language Processing (NLP), setting new benchmarks in various applications \citep{vaswani2017attention, devlin-etal-2019-bert, Radford2019LanguageMA}. However, alongside these achievements, concerns have emerged about the extent to which these models memorise their training data rather than genuinely understanding and generalising linguistic patterns \citep{khandelwal2019generalization, tanzer2021memorisation}.
Memorisation in LLMs poses serious privacy and security risks, where models have been shown to reproduce verbatim passages from their training data, including sensitive personal information and copyrighted materials \citep{patilcan}. This not only presents ethical  challenges and potential legal issues, but can potentially undermine user consent when deploying models in a generative environment. Training data extraction attacks \cite{carlini2021extracting} demonstrate that adversaries can recover spans of pretraining sample data, highlighting the practical threat of generative model deployment.

Most existing mitigation efforts focus on unlearning strategies and regularisation techniques applied during pre-training \citep{carlini2023quantifying, cheng2021mitigating}. While valuable, these approaches often lack scalability and are not easily deployable in practice, especially given the immense computational resources required to retrain large models or apply differential privacy methods \citep{anil-etal-2022-large}. Moreover, on large datasets, exhaustive extraction tests are infeasible, making it challenging to assess and mitigate memorisation effectively. Fine-tuning pre-trained LLMs on domain-specific and instruction-specific data is a common practice to adapt models to new domains and tasks, often utilising datasets with private and sensitive information. Despite this widespread application, there is a gap in understanding how fine-tuning for domain adaptation or instruction tuning impacts memorisation dynamics. 

\begin{figure}[t!]
    \centering
    \includegraphics[width=0.3\textwidth]{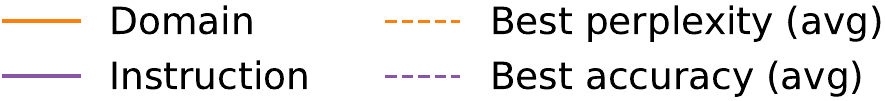}
    \begin{subfigure}[t]{0.245\textwidth}
        \includegraphics[width=\textwidth]{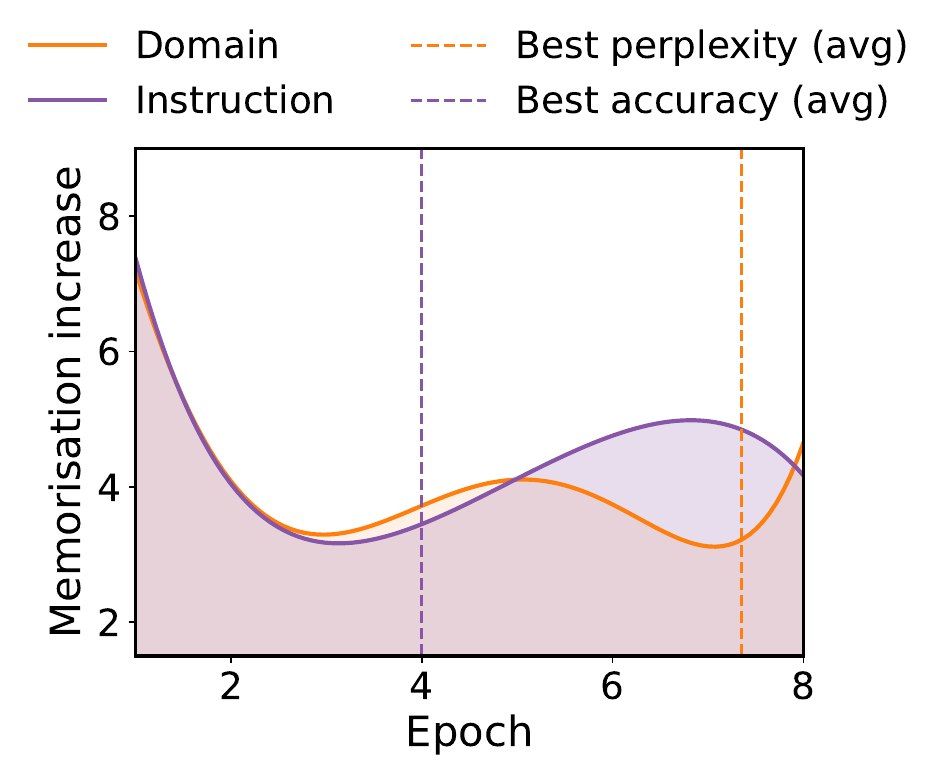}
        \caption{Pythia 1.4B}
        \label{fig:1a}
     \end{subfigure}
     \begin{subfigure}[t]{0.23\textwidth}
        \includegraphics[width=\textwidth]{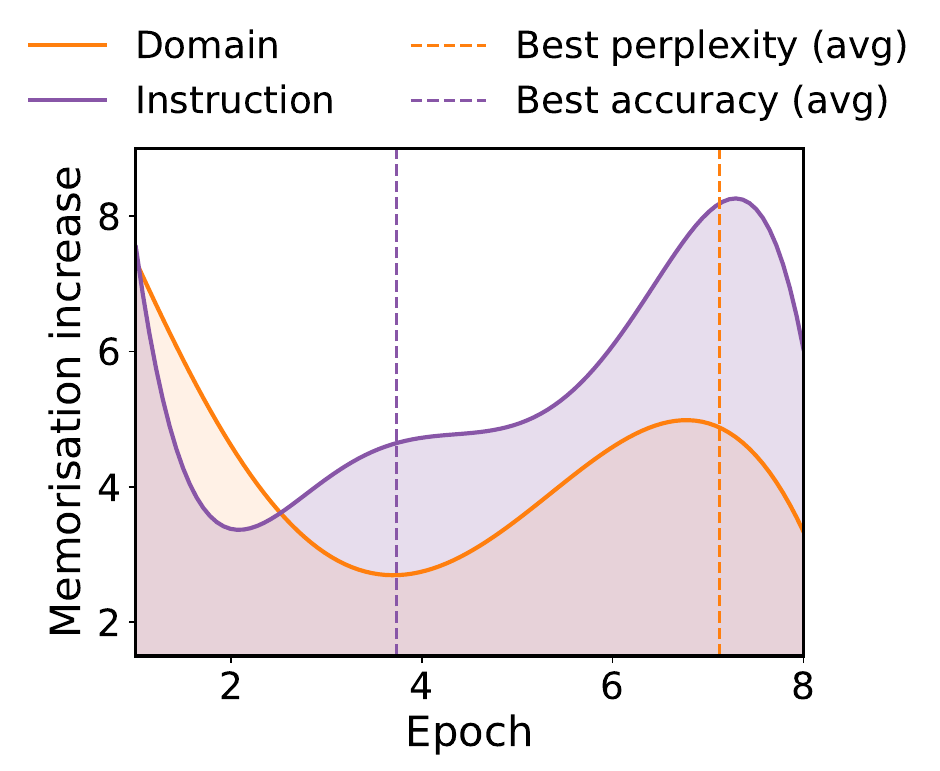}
        \caption{Pythia 12B}
        \label{fig:1b}
     \end{subfigure}
     \caption{Memorisation increases (number of new samples memorised at a given epoch) at successive fine-tuning epochs, comparing fine-tuning for domain adaptation (orange) and instruction tuning (purple) on the same data. Dashed vertical lines mark the average epoch for which validation perplexity (orange) and task evaluation accuracy (purple) are achieved, showing high memorisation before for both (a) Pythia 1.4B and (b) Pythia 12B models.}
     \label{fig:1}
\end{figure}

Our preliminary observations, illustrated in Fig. \ref{fig:1}, show significant memorisation occurring early during fine-tuning, before the model achieves optimal validation perplexity or task evaluation performance. This suggests that LLMs rapidly memorise new information before reaching typical early stopping criteria, potentially exposing sensitive information.
Owing to this, we conduct an empirical investigation into memorisation in LLMs during fine-tuning, focusing on fast, deployable mitigation strategies and insights applicable during both domain adaptation and instruction tuning, leveraging widely used memorisation metrics. We perform fine-tuning experiments with the \textit{Pythia} \citep{biderman2023pythia} models family across multiple parameter scales (1.4B - 12B), as well as \textit{Llama2} 7B \citep{touvron2023llama}, \textit{Llama3} 8B and 70B \citep{grattafiori2024llama},
and \textit{Mistral} 7B 
\citep{jiang2023mistral} models for both domain adaptation and instruction tuning across a range of common LLM evaluation datasets.

Our key contributions are:

\begin{itemize}
\item \textbf{Memorisation dynamics during fine-tuning paradigms}: we examine how memorisation manifests during common fine-tuning approaches; domain adaptation and instruction tuning, across a wide range of both model sizes and datasets. 

\item \textbf{N-gram memorisation as a precursor to verbatim memorisation}: we use an $n$-gram based partial memorisation metric as an early indicator of longer phrase memorisation, finding high-rates in samples prior to becoming memorised, across the majority of datasets, model scales, and fine-tuning methods.

\item \textbf{Optimal stopping criteria}: we identify optimal stopping criteria during fine-tuning that significantly reduce memorisation with minimal impact on performance, providing a generalisable heuristic to mitigate memorisation risks in real time.

\item \textbf{Comparison of mitigation techniques}: we explore loss-based regularisation approaches, demonstrating further reductions in memorisation that are scalable, generalisable, and competitive with an existing approach.

\end{itemize}

\begin{table*}[t]
\centering
\resizebox{0.98\textwidth}{!}{
\begin{tabular}{lllll}
\toprule
\textbf{Category} & \textbf{Dataset} & \textbf{Domain} & \textbf{Input type} & \textbf{Output type} \\
\midrule
Classification & SST-5 & Movie reviews & Single review sentence & 5-way sentiment \\
               & QQP & Quora community QA & Question\textsubscript{1}, Question\textsubscript{2} & Duplicate? (yes/no) \\
               & RTE & News \& Wikipedia & Premise, Hypothesis & Entail / Not-entail \\
               & WANLI & MultiNLI-derived genres & Premise, Hypothesis & Entail / Neutral / Contradict \\
\addlinespace
Question-Answering (QA) & SQuAD v2 & Wikipedia articles & Paragraph context, Question & Answer span or \texttt{[NoAnswer]} \\
                   & HellaSwag & WikiHow narratives & Context + 4 candidate endings & Correct ending (MC-4) \\
                   & PubMedQA & Biomedical abstracts & Abstract (no conclusion), Question & Yes / No / Maybe \\
\addlinespace
Summarisation & XSum & BBC news & Full news article & One-sentence abstractive summary \\
              & CNN/DailyMail & CNN \& Daily Mail news & Full news article & Multi-sentence highlights \\
\addlinespace
Instruction Following & Alpaca & Mixed user prompts & Instruction (± optional input) & Free-form response \\
                      & FLAN v2 & Multi-domain tasks & Instruction template & Free-form response \\
\bottomrule
\end{tabular}}
\caption{Summary and grouping of datasets used for performing fine-tuning.}
\label{tab:1}
\end{table*}

\section{Related Work}
Prior research on memorisation in LLMs spans three main areas: measurement, characterisation across pre-training versus fine-tuning, and mitigation, each covered in the following section.

\subsection{Measuring Memorisation}
Evaluating the extent of memorisation in LLMs necessitates robust metrics and evaluation techniques. \citet{carlini2023quantifying} introduce the concept of \emph{k-extractable memorisation}, which measures a model's tendency to reproduce training data when provided with specific input prefixes, representing a stringent test for data leakage. Complementary approaches include membership inference attacks aimed at classifying pretraining samples \citep{shokri2017membership}. Memorisation and generalisation have been shown to carry some interdependent relationships \cite{tanzer2021memorisation, khandelwal2019generalization, yeom2018privacy}, with memorisation dynamics in large scale LLMs studied in \citet{tirumala2022memorization, carlini2023quantifying}.

\subsection{Memorisation in Pre-training Versus Fine-tuning}
The dynamics of memorisation exhibit distinct characteristics during the pre-training and fine-tuning stages of LLM development. In the pre-training phase, models are exposed to extensive and often publicly available datasets, where factors such as data redundancy and model size play critical roles in determining the extent of memorisation \citep{tanzer2021memorisation, khandelwal2019generalization, carlini2023quantifying, carlini2019secret}. Research indicates that larger models are more prone to rapidly memorising training data \citep{tirumala2022memorization, nasr2023scalable}. Conversely, during fine-tuning on specialized or private datasets, different memorisation risks emerge. Studies have demonstrated that specific fine-tuning methodologies, like adapter-based techniques, can reduce the likelihood of memorising sensitive information \citep{mireshghallah2022memorization, dodge2021documenting, raffel2020exploring}. Additionally, counterfactual memorisation assessments \citep{zhang2021counterfactual} aid in distinguishing between memorisation arising from pre-training and that from fine-tuning, thereby informing targeted mitigation strategies tailored to each training phase.

\subsection{Mitigation Strategies and Regularisation}
During the training process, regularisation methods such as the addition of noise to input embeddings \citep{jain2024neftune} are employed to mitigate memorisation \citep{feldman2020neural, tirumala2022memorization}. Post-training techniques include fine-tuning and machine unlearning approaches \citep{maini2023can}, which aim to remove specific data from the model without necessitating a complete retraining. Despite these measures, achieving a balance between preserving model performance and ensuring data privacy remains a significant challenge.
Mitigating memorisation in language models is critical for preserving privacy and preventing the leakage of sensitive information. Conventional regularisation techniques, such as weight decay and dropout, are designed to prevent overfitting and thereby reduce memorisation \citep{feldman2020neural}. However, these methods have proven inadequate in fully reducing memorisation within LLMs \citep{tirumala2022memorization}. Advanced regularisation approaches, including data-dependent token dropout \citep{hans2024like} and targeted token masking \citep{jain2024neftune}, offer partial mitigation but often fail to eliminate the risk of memorising entire data passages, especially when dealing with highly duplicated datasets.

\section{Methodology}
We begin by defining how we will measure memorisation, leveraging an existing approach and introducing a partial measure for fine-grained measurement. We follow this by introducing the experimental setup of our study for fine-tuning for domain adaptation and instruction tuning.

\subsection{Memorisation Metrics}
For an exact and scalable measure of verbatim memorisation, we employ the widely used extraction metric introduced in \citet{carlini2023quantifying}.\\

\noindent \textbf{Memorisation:} Let $f$ be a generative LLM trained on data $D$, with prefix-suffix pair $(p, s)$ contained within a sample in $D$. A suffix $s$ is said to be \emph{$k$-extractable} (\emph{memorised}) if $f$ generates a string containing $s$ exactly when prompted with a prefix of length $k$ using greedy decoding.

Therefore we can compute the percentage of the fine-tuning data memorised at each fine-tuning epoch as: 
\[
\text{Mem} = \left( \frac{\sum k\text{-extractable suffixes } s}{\text{total samples in data }D} \right) \times 100.
\]
This definition provides a directly computable metric on the generated output from our fine-tuned models, allowing fast evaluation at each fine-tuning epoch. We use the above as the definition for \emph{memorisation} throughout.

\subsection{N-gram Memorisation}

For fine-grained measure of memorisation, we implement a partial memorisation metric.\\

\noindent \textbf{N-gram Memorisation:} For a set of $n$-gram sizes \( N = \{n_1, n_2, \dots, n_k\} \), the $n$-gram memorisation score between the model's output \( f(p) \) and the target sequence \( s\) is defined as the proportion of matching $n$-grams of sizes in \( N \), where the matches are exact for each $n$-gram but invariant to ordering of $n$-grams.

Formally, given $M_i$ as the fraction of matching $n$-grams in $N$ between \( f(p_i) \) and \( s_i \), the $n$-gram memorisation score is then calculated as:
\[
    \text{$n$-gram Mem} = \left(\frac{\sum_{d \in D} M_d}{|D|}\right) \times 100.
\]

This metric provides a finer-grained measure of memorisation that allows for different lengths and number of $n$-grams, which can be tuned for suitability for specific datasets, sequence length sizes, and granularity of sensitive information.

\subsection{Datasets}
\label{sec:datasets}

\begin{figure*}[t!]
    \centering
     \includegraphics[width=0.9\textwidth]{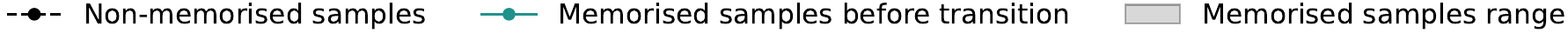}\\
    \vspace{3pt}
    \begin{subfigure}[t]{0.25\textwidth}
        \includegraphics[width=\textwidth]{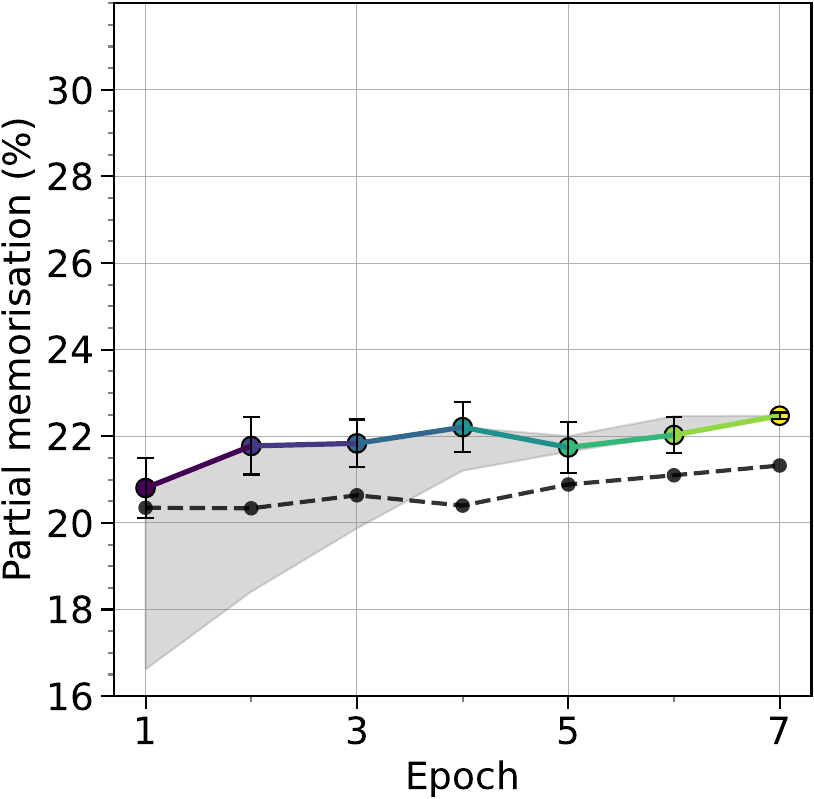}
        \caption{Summarisation}
        \label{fig:2a}
     \end{subfigure}
     \begin{subfigure}[t]{0.235\textwidth}
        \includegraphics[width=\textwidth]{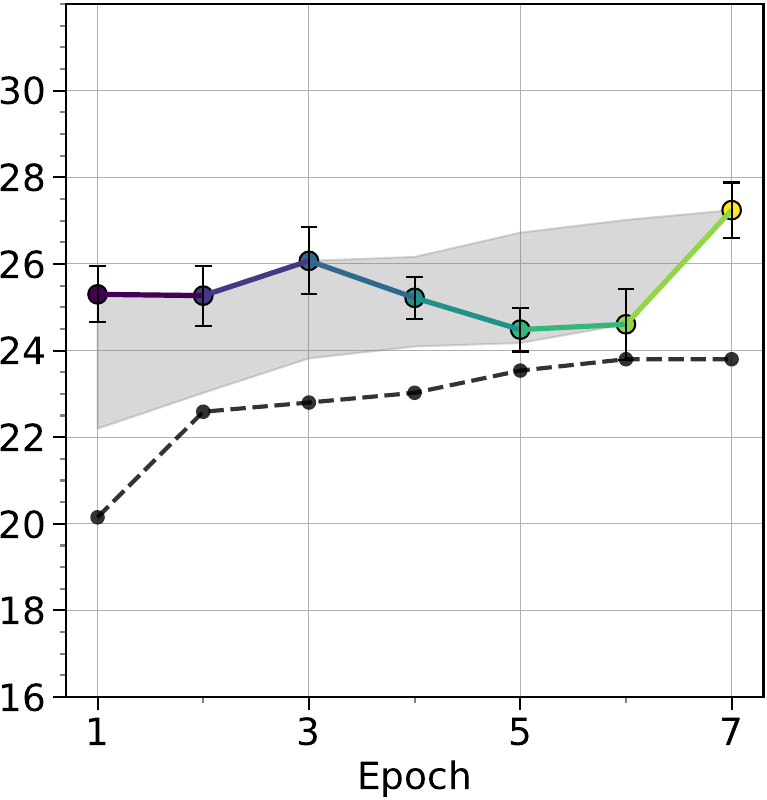}
        \caption{Summarisation (domain)}
        \label{fig:2b}
     \end{subfigure}
     \begin{subfigure}[t]{0.235\textwidth}
        \includegraphics[width=\textwidth]{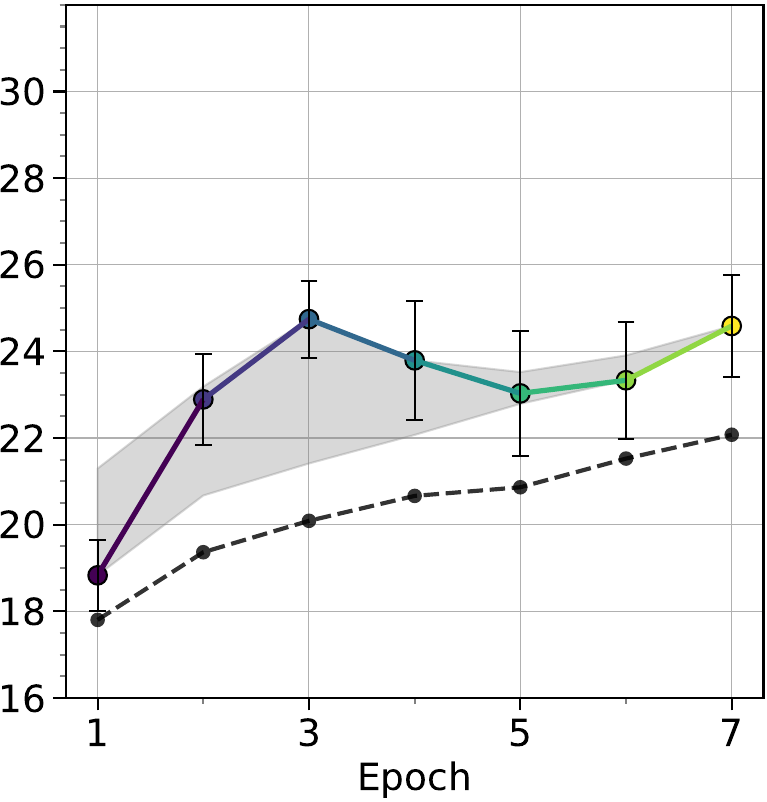}
        \caption{Instruction}
        \label{fig:2c}
     \end{subfigure}
     \begin{subfigure}[t]{0.261\textwidth}
        \includegraphics[width=\textwidth]{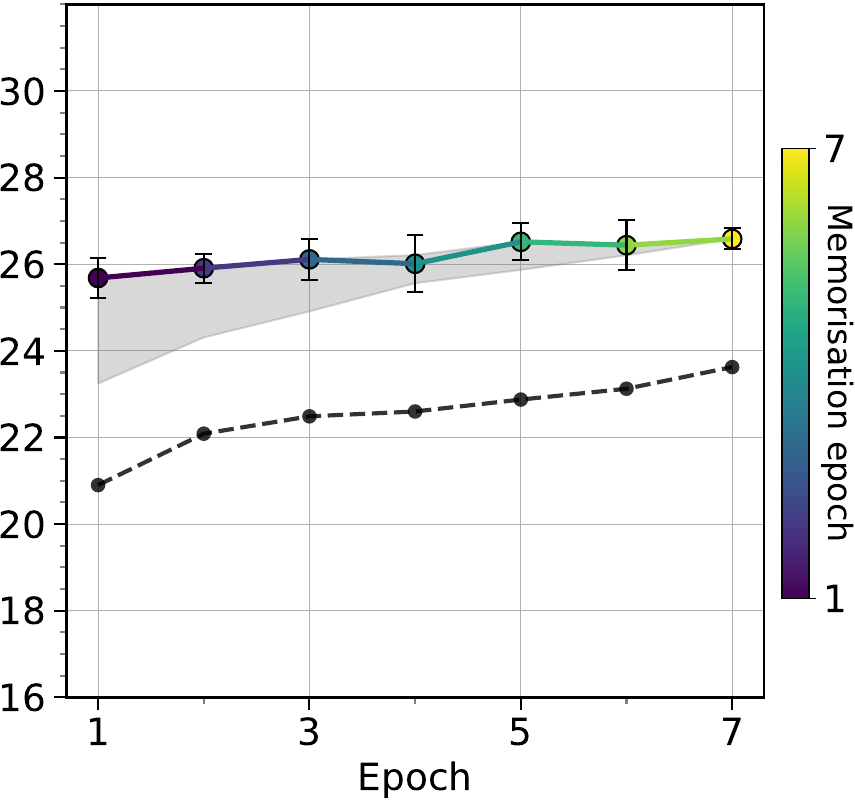}
        \caption{Instruction (domain)}
        \label{fig:2d}
     \end{subfigure}\\
     \vspace{5pt}
     \begin{subfigure}[t]{0.25\textwidth}
        \includegraphics[width=\textwidth]{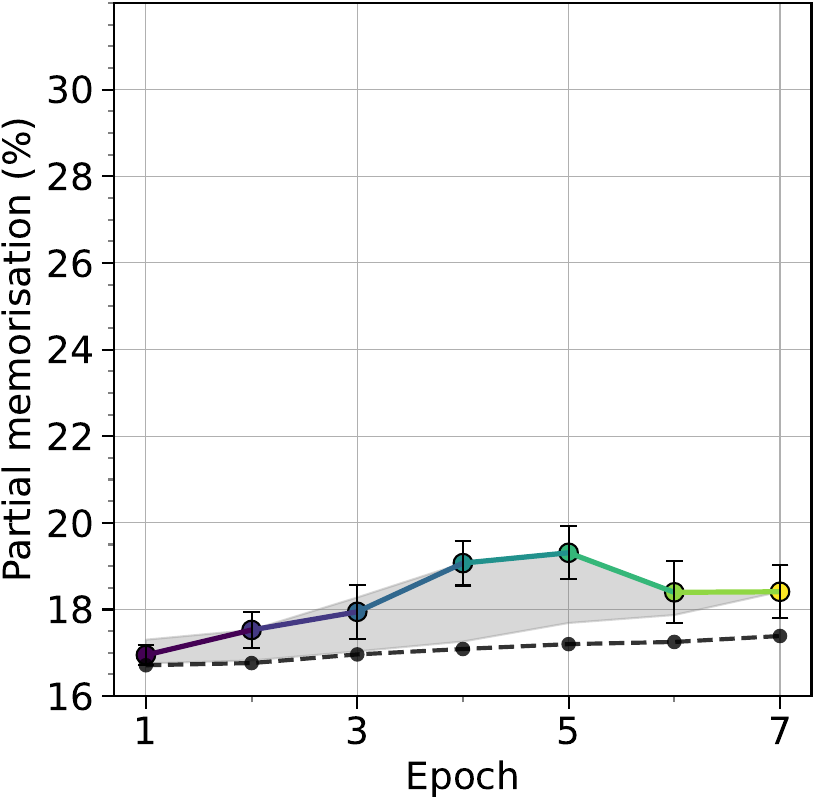}
        \caption{\small{QA}}
        \label{fig:2e}
     \end{subfigure}
     \begin{subfigure}[t]{0.235\textwidth}
        \includegraphics[width=\textwidth]{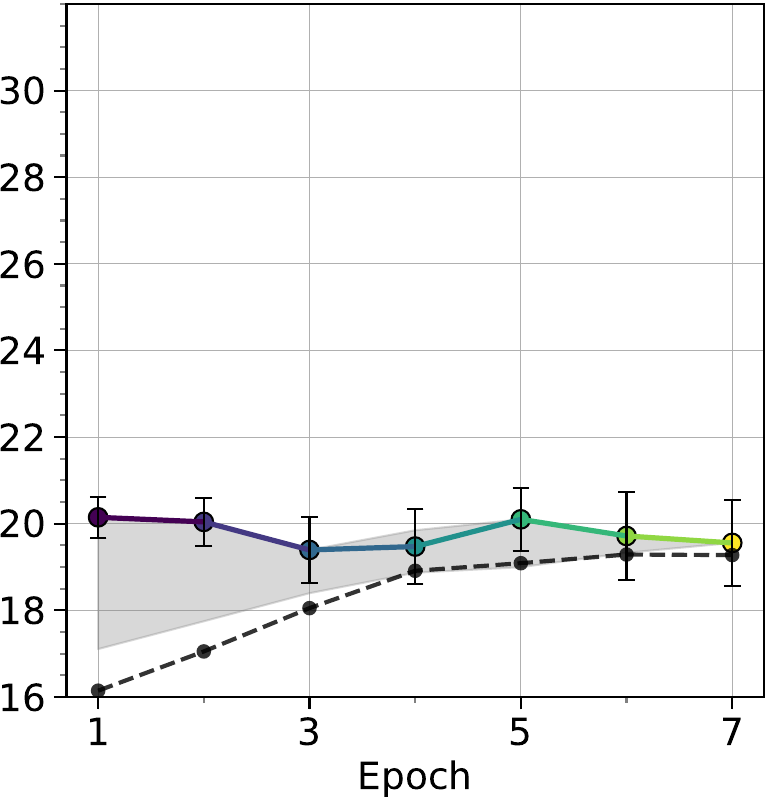}
        \caption{\small{QA (domain)}}
        \label{fig:2f}
     \end{subfigure}
     \begin{subfigure}[t]{0.235\textwidth}
        \includegraphics[width=\textwidth]{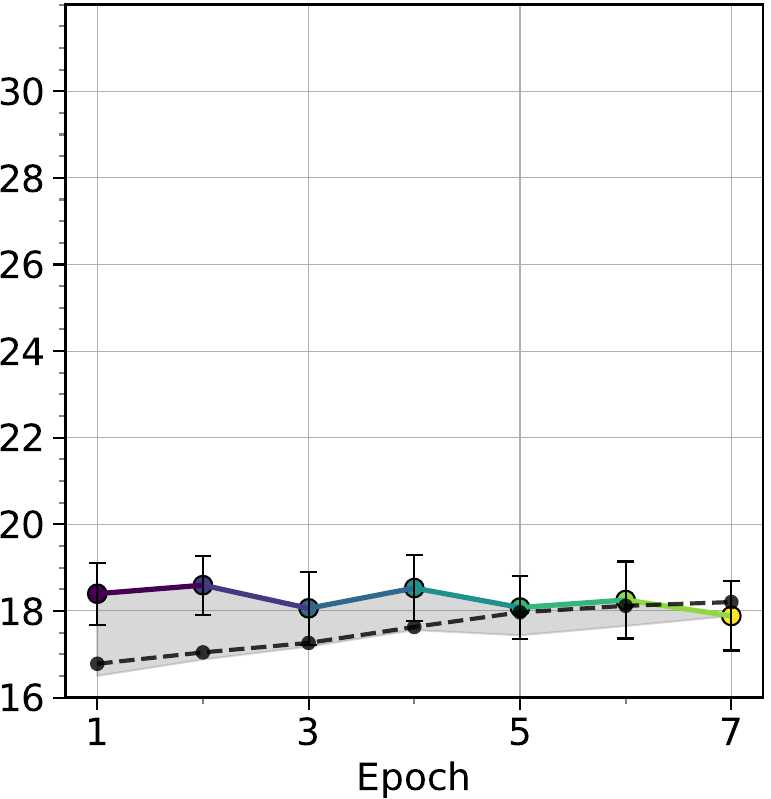}
        \caption{Classification}
        \label{fig:2g}
     \end{subfigure}
     \begin{subfigure}[t]{0.261\textwidth}
        \includegraphics[width=\textwidth]{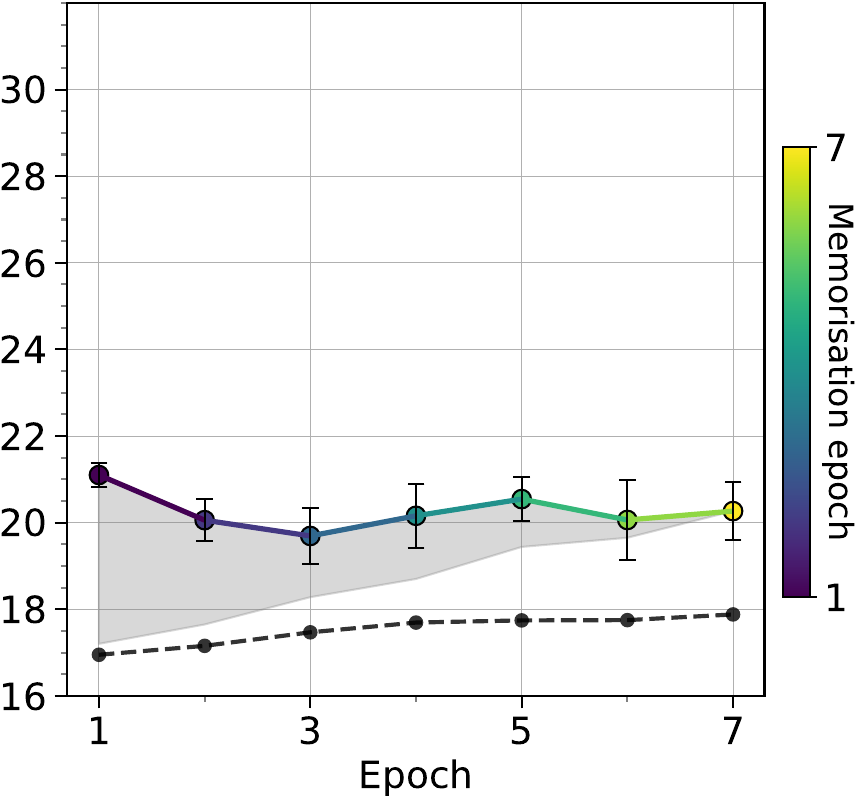}
        \caption{Classification (domain)}
        \label{fig:2h}
     \end{subfigure}
     \caption{Partial \(n\)-gram memorisation across fine-tuning epochs for the four dataset categories, with \textit{domain} indicating domain adaptation fine-tuning. In each panel the coloured solid line reports, at epoch \(t\), the \emph{median} score of samples that become memorised at subsequent epoch \(t{+}1\); the point colour encoding that memorisation epoch. The grey shaded region spans the full score range of \emph{all} samples that ever become memorised, irrespective of when the transition occurs. Error bars show the standard deviation over five random seeds, while the black dashed line is the baseline for samples that are never memorised. Results are averages over Pythia model sizes from 1.4 B to 12 B parameters.}
    \label{fig:2}
\end{figure*}

We leverage datasets taken from three open instruction pools—the Public Pool of Prompts (\textbf{P3}) \citep{sanh2021multitask}, the \textbf{FLAN} collection \citep{wei2023flan}, and the \textbf{Alpaca-52K} corpus \citep{taori2023alpaca}. We conduct both instruction tuning and domain adaptation experiments by choosing to include/remove the task-specific instruction prompt for each dataset. These datasets encompass a range of core NLP task types—including classification, Natural Language Inference (NLI), coreference resolution, Question-Answering (QA), and free‐form instruction following—and span diverse domains such as encyclopedic, news, clinical notes, biomedical research, and social media text. A summary of all datasets used is outlined in Table \ref{tab:1}, with further details in Appendix \ref{section:a1}.
We categorise them into the following:

\begin{itemize}
    \item \textbf{Classification \& NLI}: short, label-based prompts (sentiment, paraphrase, entailment). 
    \item \textbf{Question-Answering}: a mix of extractive, multiple-choice, and yes/no items.
    \item \textbf{Summarisation}: single-sentence (XSum) and multi-sentence (CNN/DailyMail) summaries. 
    \item \textbf{Instruction Following}: open-ended prompts from Alpaca and FLAN tasks. 
\end{itemize}

\begin{figure*}[t!]
    \centering
     \includegraphics[width=0.9\textwidth]{figures_final/partial/partial_legend.pdf}\\
    \vspace{3pt}
    \begin{subfigure}[t]{0.25\textwidth}
        \includegraphics[width=\textwidth]{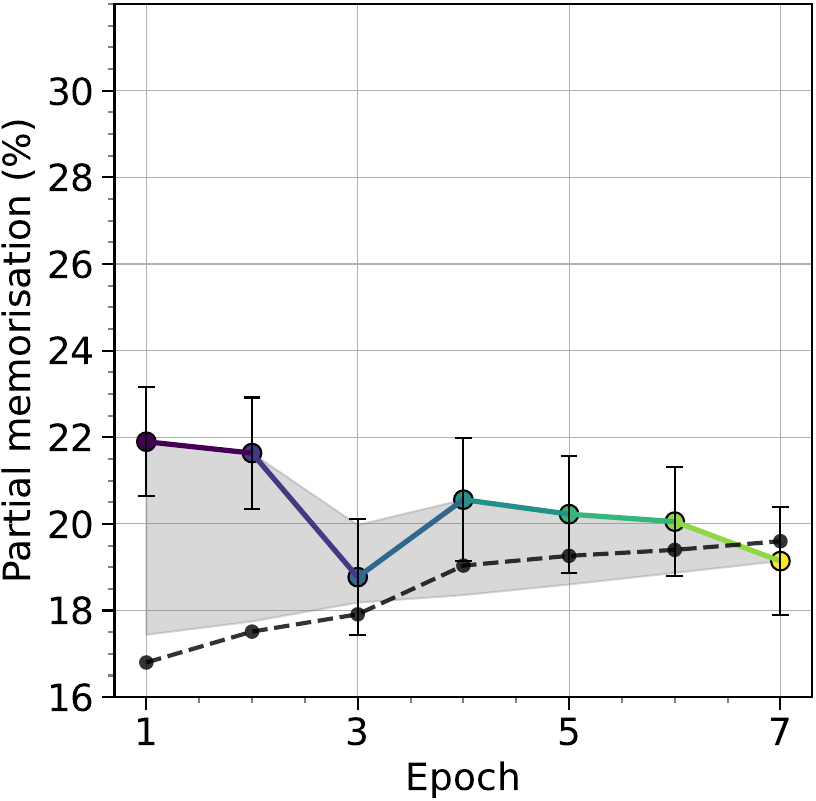}
        \caption{Pythia 1.4B}
        \label{fig:3a}
     \end{subfigure}
     % \hspace{1pt}
     \begin{subfigure}[t]{0.235\textwidth}
        \includegraphics[width=\textwidth]{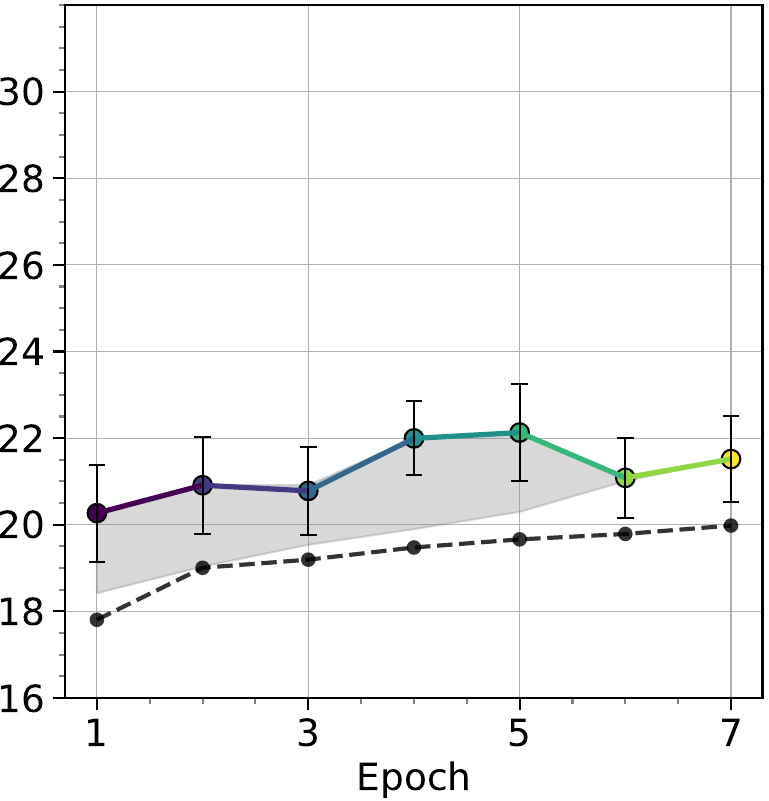}
        \caption{Pythia 2.8B}
        \label{fig:3b}
     \end{subfigure}
     % \hspace{1pt}
     \begin{subfigure}[t]{0.235\textwidth}
        \includegraphics[width=\textwidth]{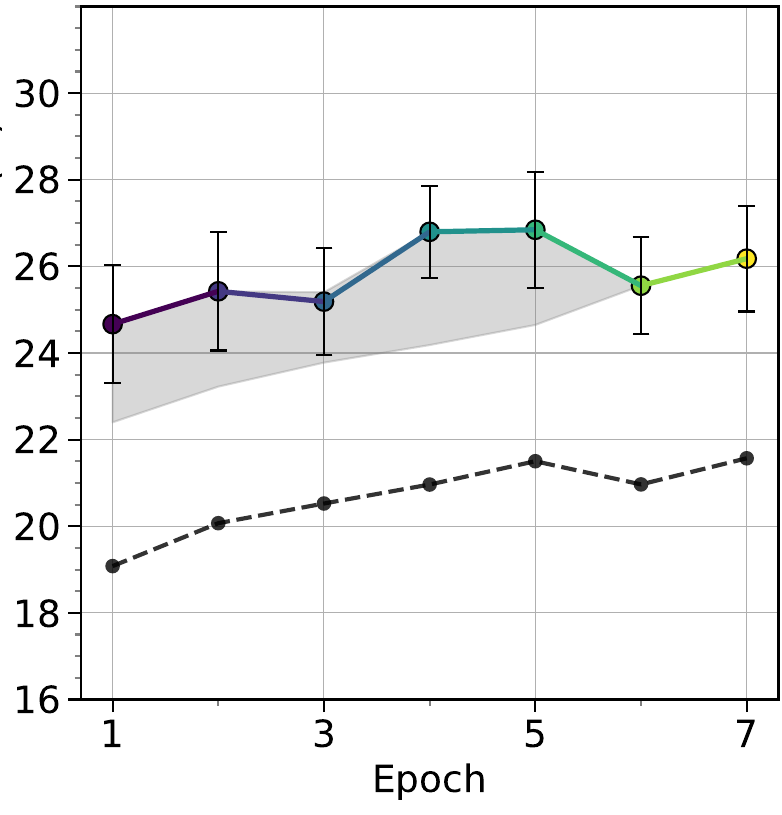}
        \caption{Pythia 6.9B}
        \label{fig:3c}
     \end{subfigure}
     % \hspace{1pt}
     \begin{subfigure}[t]{0.26\textwidth}
        \includegraphics[width=\textwidth]{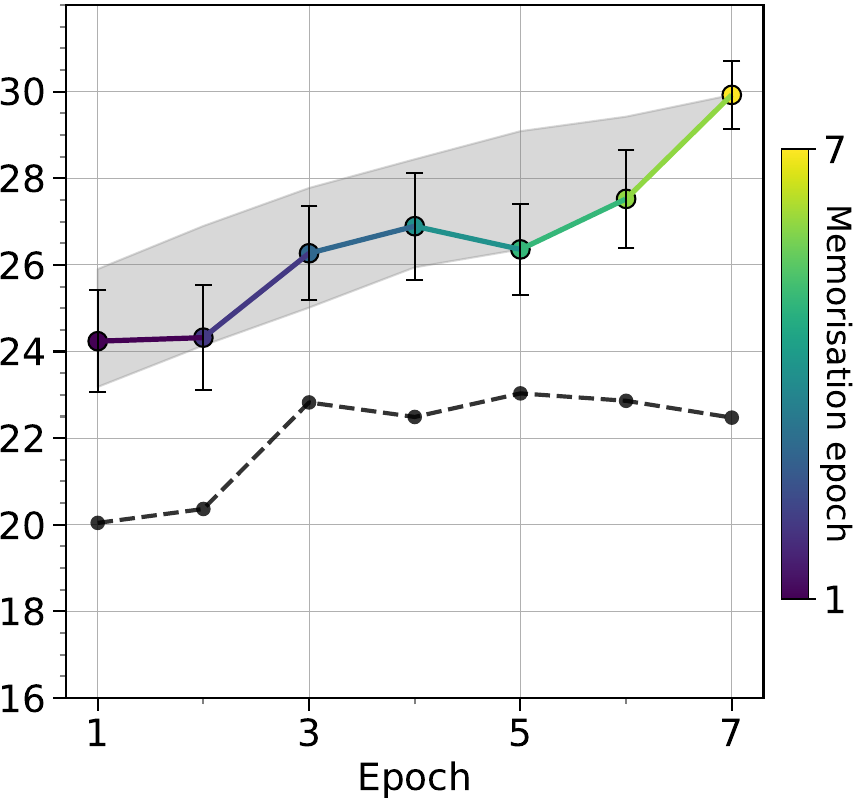}
        \caption{Pythia 12B}
        \label{fig:3d}
     \end{subfigure}\\
     \vspace{5pt}
     \begin{subfigure}[t]{0.25\textwidth}
        \includegraphics[width=\textwidth]{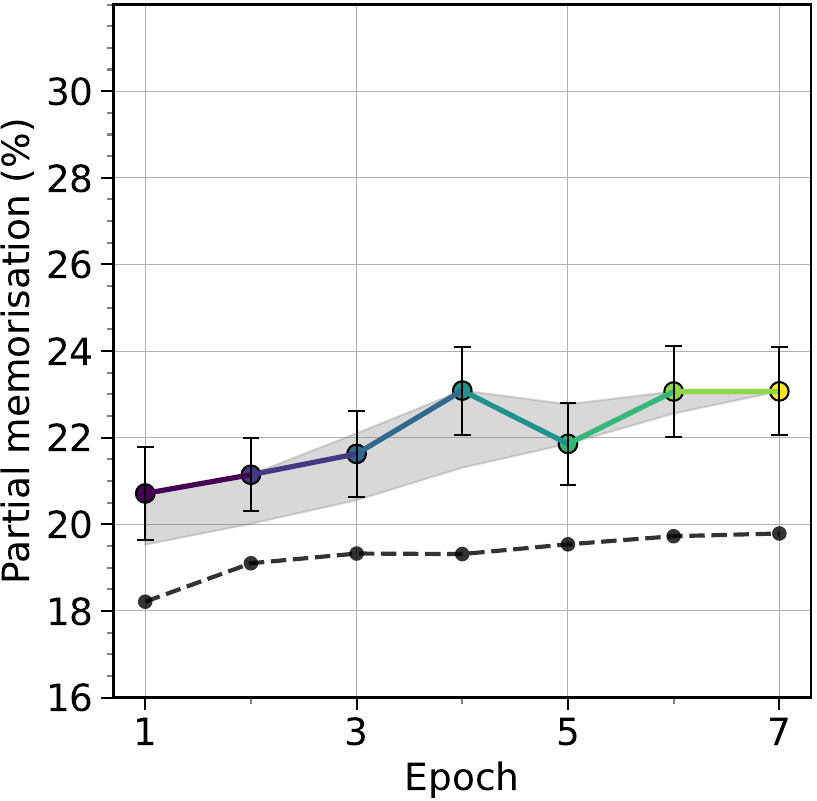}
        \caption{Mistral 7B}
        \label{fig:3e}
     \end{subfigure}
     % \hspace{1pt}
     \begin{subfigure}[t]{0.235\textwidth}
        \includegraphics[width=\textwidth]{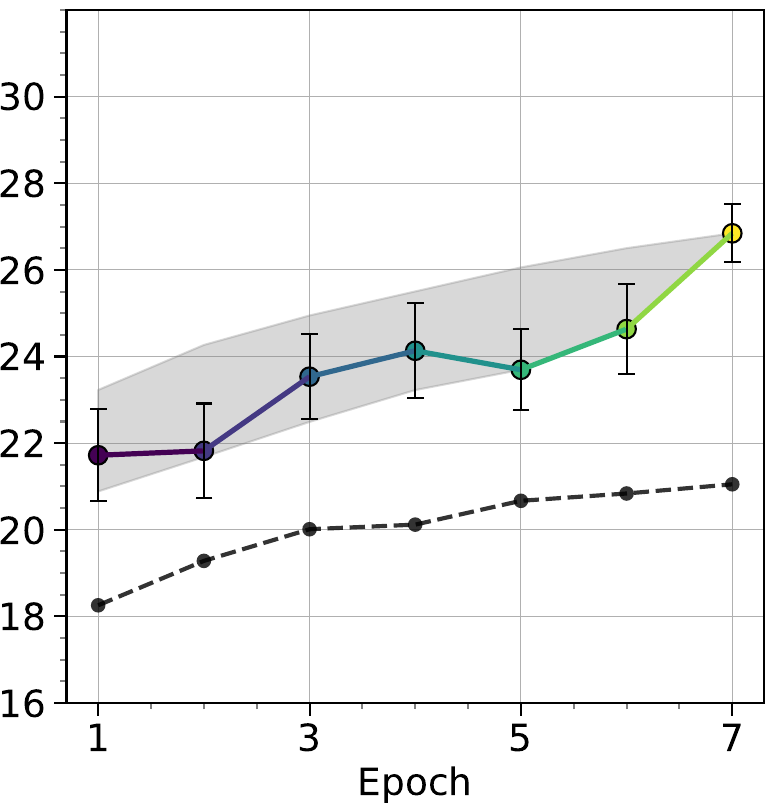}
        \caption{Llama3 8B}
        \label{fig:3f}
     \end{subfigure}
     % \hspace{1pt}
     \begin{subfigure}[t]{0.26\textwidth}
        \includegraphics[width=\textwidth]{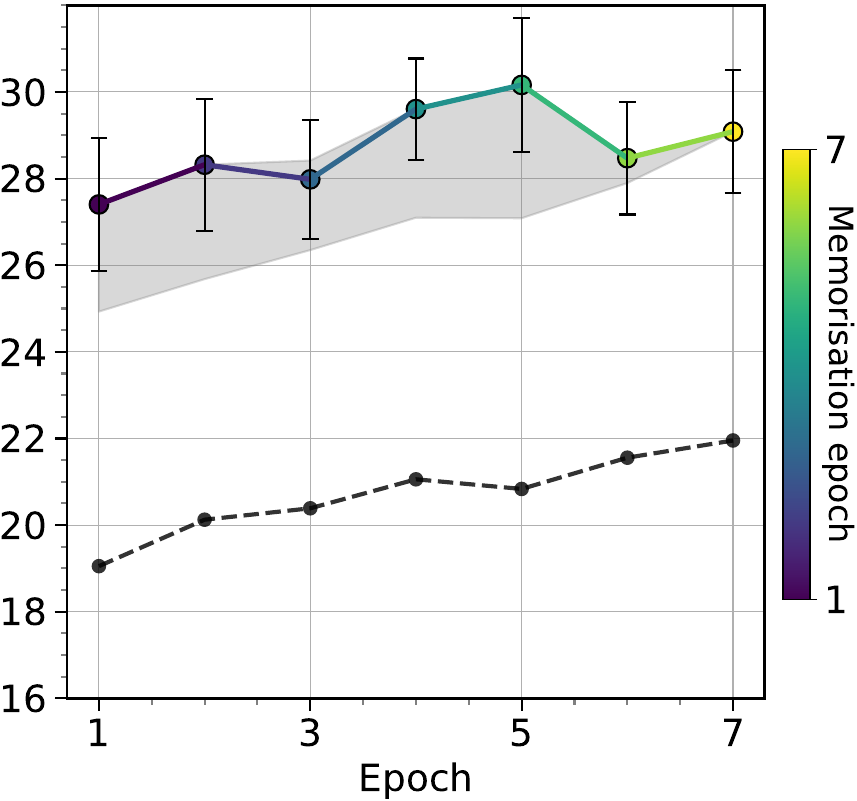}
        \caption{Llama3 70B}
        \label{fig:3g}
     \end{subfigure}
    \caption{Partial memorisation across epochs for different models. The coloured solid line gives the median score of samples that will be memorised at epoch~\(t{+}1\); point colour marks that future epoch.  The grey region shows the full range for all eventually memorised samples, while the black dashed line is the baseline for samples never memorised.  Error bars denote the standard deviation across five random seeds.}
    \label{fig:3}
\end{figure*}

These datasets are chosen to provide task and domain diversity for evaluating how this impacts memorisation, as well as providing datasets which can be used for both domain adaptation and instruction tuning. 

\subsection{Pre-trained Models}
\label{sec:models}

Experiments are run on the \textit{Pythia} model family \citep{biderman2023pythia} using sizes of 1.4B, 2.8B, 6.9B, and 12B parameters. Pythia pre-training keeps hyper-parameters and dataset composition fixed while doubling model size at each step, giving a clean scaling ladder to evaluate on. Additionally, we use \textit{Llama2} 7B \citep{touvron2023llama}, \textit{Llama3} 8B and 70B \citep{grattafiori2024llama},
and the \textit{Mistral 7B} model
\citep{jiang2023mistral}. These models are chosen to enable comparisons between architectural variants at similar model sizes. All pre-trained model checkpoints are publicly accessible via Huggingface \citep{wolf2019huggingface}. Fine-tuning is performed using the Adam optimiser \cite{kingma2014adam}. We perform full‑parameter fine‑tuning and, for comparison, perform \textit{partial fine‑tuning} in which only the top $n$ transformer layers are updated while the rest remain frozen, enabling us to measure how restricting the trainable subset of parameters alters memorisation behaviour.

\subsection{Fine-Tuning Approach} We employ domain adaptation and instruction tuning by fine-tuning each model for up to 8 epochs on a maximum of 5,000 samples of the target dataset. When performing domain adaptation, we simply remove the task-specific instructions from the input. We evaluate on a held-out validation set for both \textit{validation perplexity} and task-specific \textit{evaluation performance}. For evaluation performance, we use the standard evaluation metrics used to measure performance for that task (details of each can be found in the Appendix \ref{section:a1}). For our memorisation and $n$-gram memorisation metrics, we evaluate on the 5,000 samples of training data used to fine-tune. The small number of fine-tuning samples allows us to rapidly experiment over model scales and datasets while maintaining relevant to typically small and private fine-tuning datasets. Evaluations are performed at each epoch to monitor the progression of memorisation relative to validation performance and evaluation performance. 

Following \citet{carlini2023quantifying}, we test \(k\)-extractable memorisation with three prefix lengths, \(k\in\{12,16,20\}\), and a fixed 20‑token suffix. Lengths below 12 tokens collide frequently across corpora, whereas prefixes longer than 20 tokens limits the number of samples we can use from each dataset. We empirically find that 4, 5, and 6-grams for our $n$-gram memorisation metric provides a good signal for highly memorised phrases without being computationally prohibitive. All results are averaged over 10 runs with random seed initialisations. For robustness, we use different randomly sampled prefix-suffix pairs for each of the 10 randomly initialised fine-tuning runs. 

\begin{figure}[t!]
    \centering
     \includegraphics[width=0.49\textwidth]{figures_final/partial/partial_legend.pdf}\\
    \vspace{3pt}
    \begin{subfigure}[t]{0.23\textwidth}
        \includegraphics[width=\textwidth]{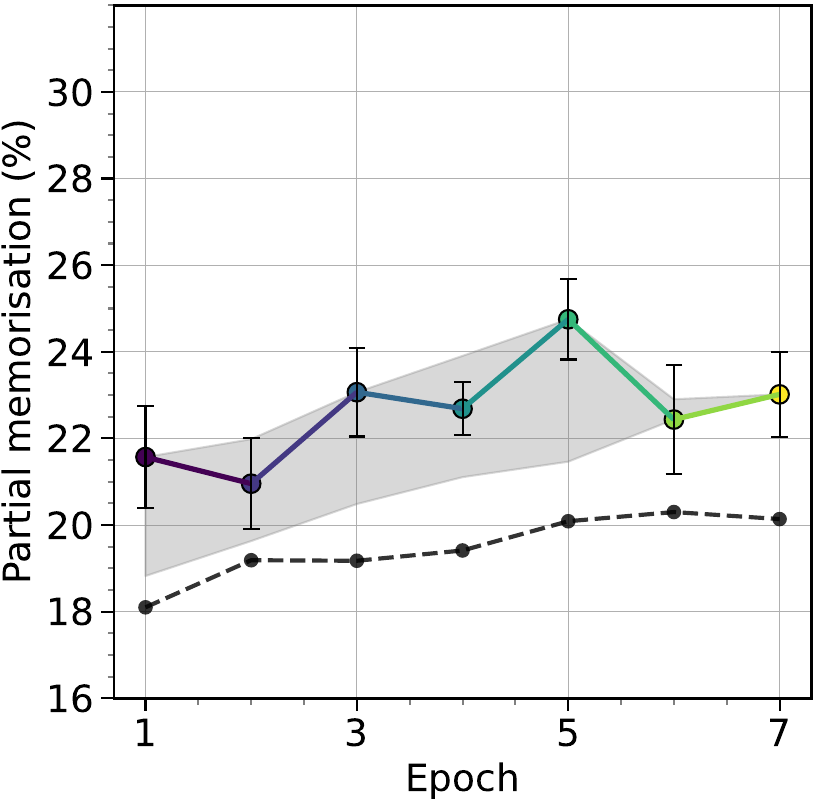}
        \caption{Final layer}
        \label{fig:4a}
     \end{subfigure}
     % \hspace{1pt}
     \begin{subfigure}[t]{0.239\textwidth}
        \includegraphics[width=\textwidth]{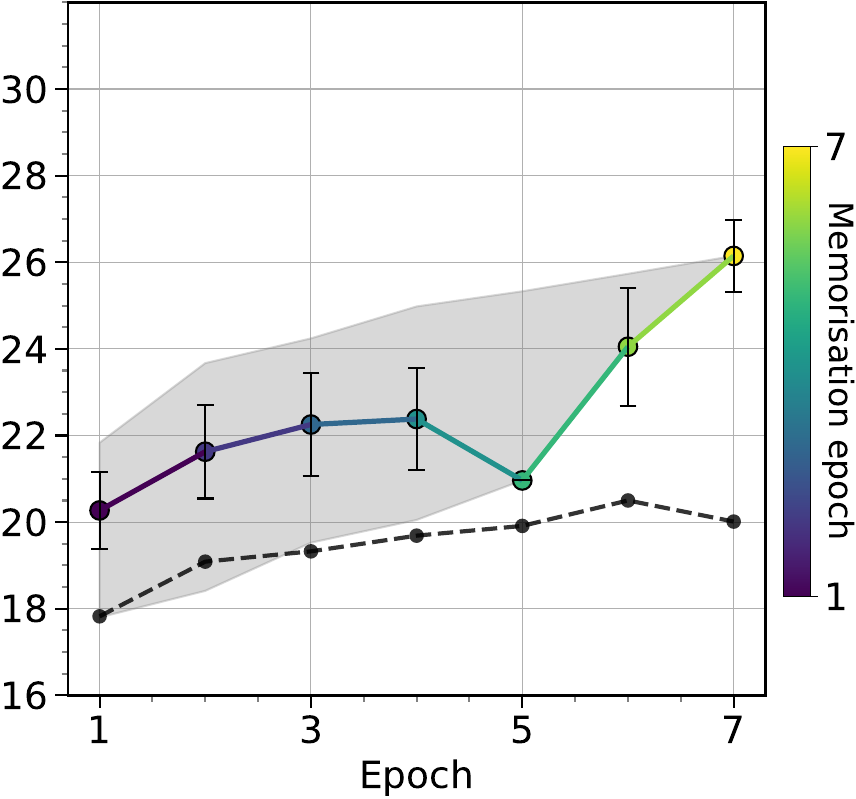}
        \caption{Final 4 layers}
        \label{fig:4b}
     \end{subfigure}\\
     % \hspace{1pt}
     \vspace{3pt}
     \begin{subfigure}[t]{0.23\textwidth}
        \includegraphics[width=\textwidth]{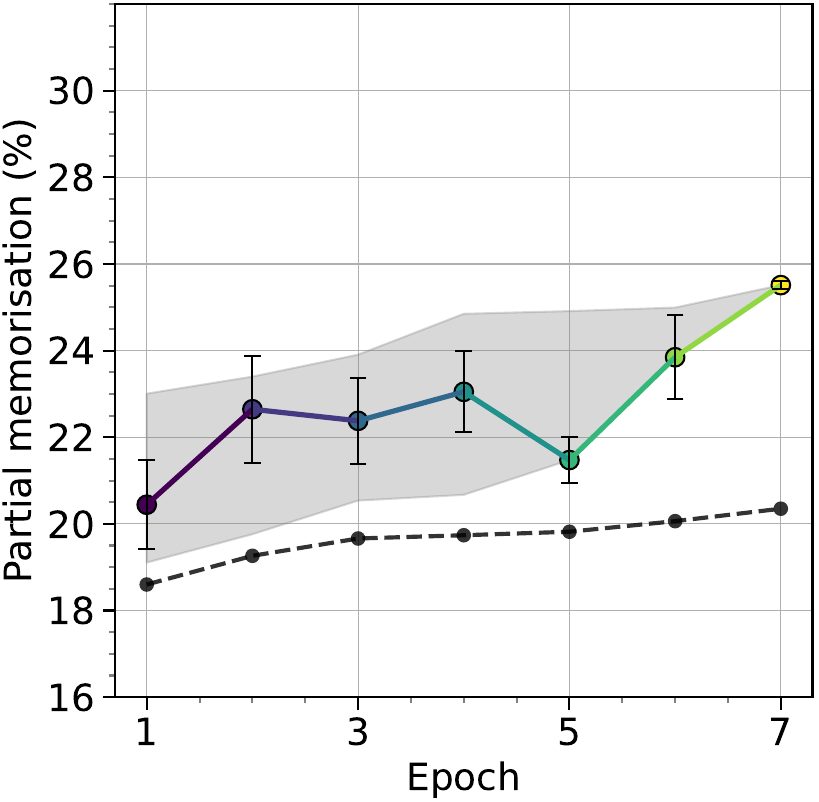}
        \caption{Final 8 layers}
        \label{fig:4c}
     \end{subfigure}
     % \hspace{1pt}
     \begin{subfigure}[t]{0.239\textwidth}
        \includegraphics[width=\textwidth]{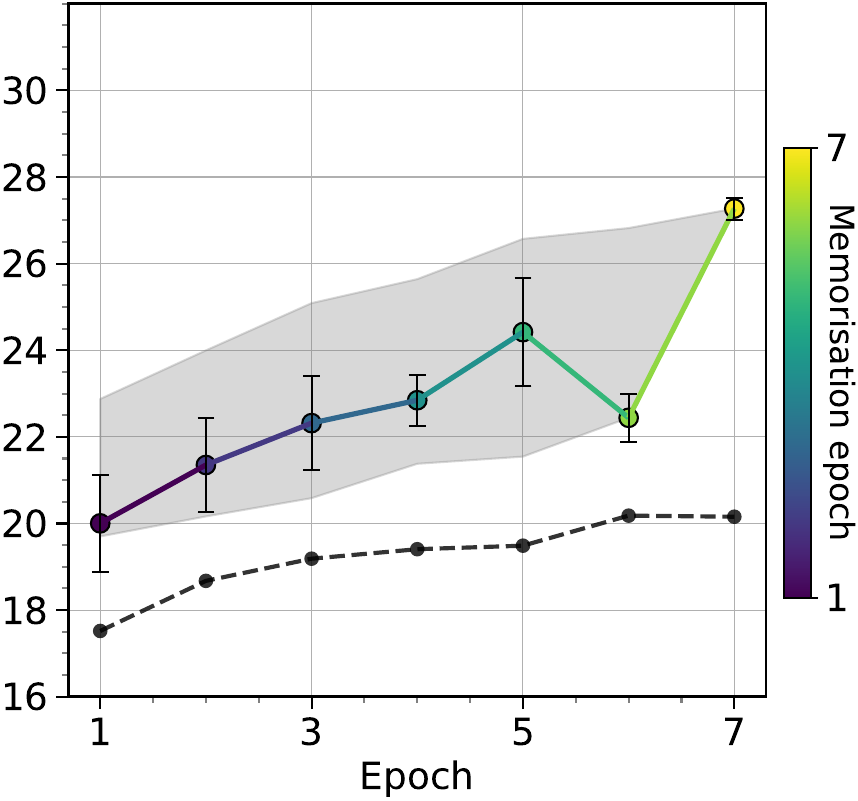}
        \caption{Final 16 layers}
        \label{fig:4d}
     \end{subfigure}
     \caption{Final-layer partial fine-tuning comparison of the Llama3 8B model. The final $n$ layers of the model are unfrozen and updated when fine-tuning, with the remaining layers frozen.}
    \label{fig:4}
\end{figure}

\section{Results and Discussion}

We begin by evaluating $n$-gram memorisation results over model scales and domains. After, we discuss epoch selection criteria for minimising memorisation and performance trade-offs. Finally, we compare mitigation strategies across model scales.

\begin{figure*}[t!]
     \centering
     \begin{subfigure}[t]{0.53\textwidth}
         \centering
         \includegraphics[width=\textwidth]{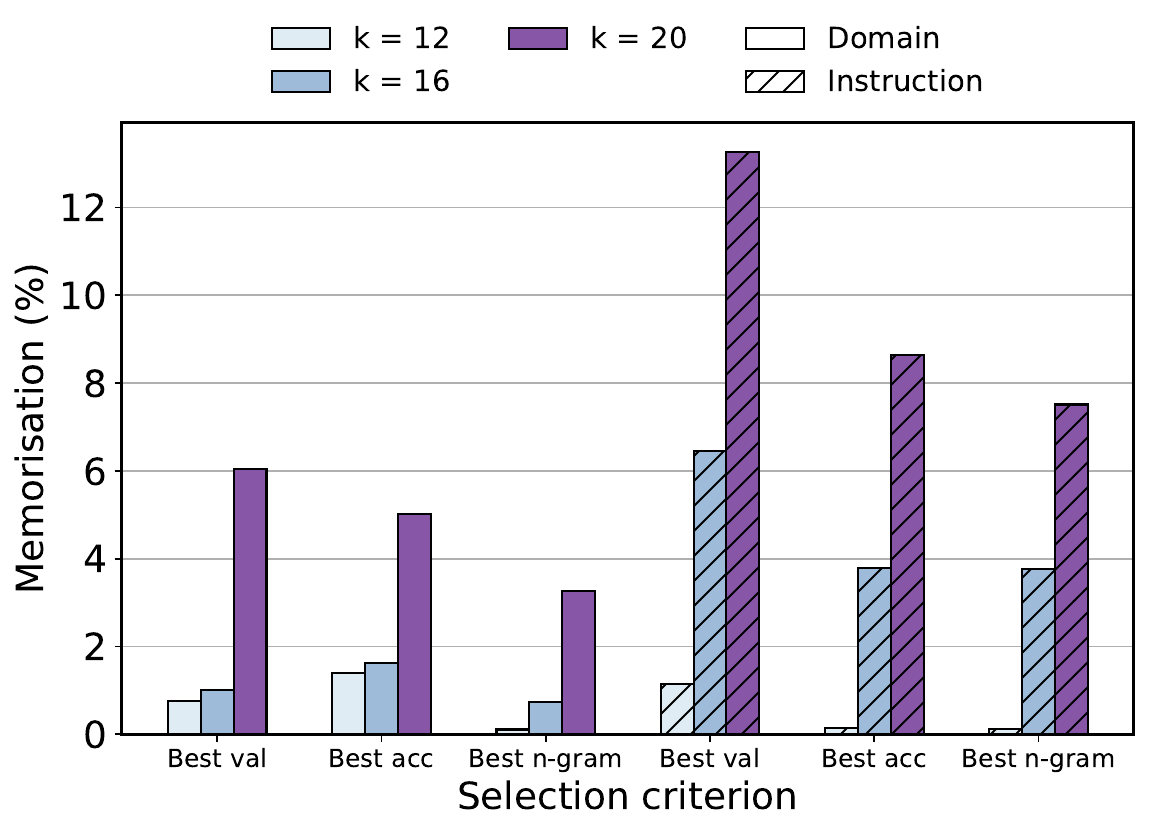}
         \caption{Pythia 1.4B}
         \label{fig:5a}
     \end{subfigure}
     \hfill
     \begin{subfigure}[t]{0.45\textwidth}
         \centering
         \includegraphics[width=\textwidth]{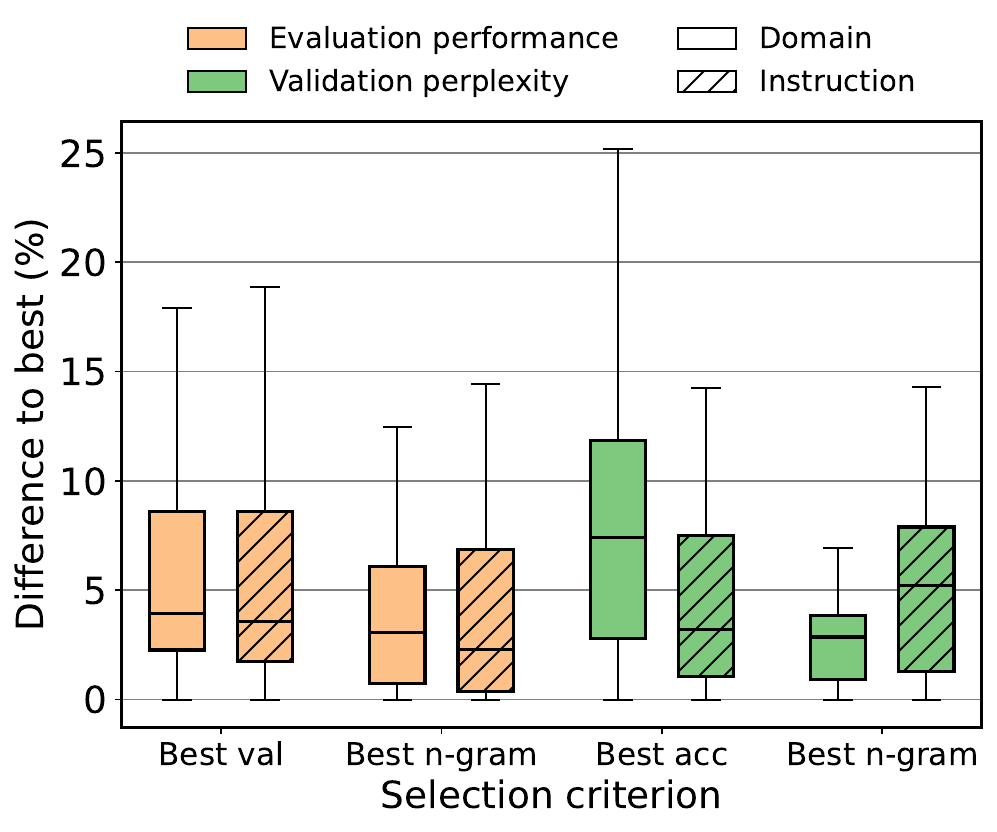}
          \caption{Llama 8B}
          \label{fig:5b}
     \end{subfigure} \\
    \begin{subfigure}[t]{0.53\textwidth}
         \centering
         \includegraphics[width=\textwidth]{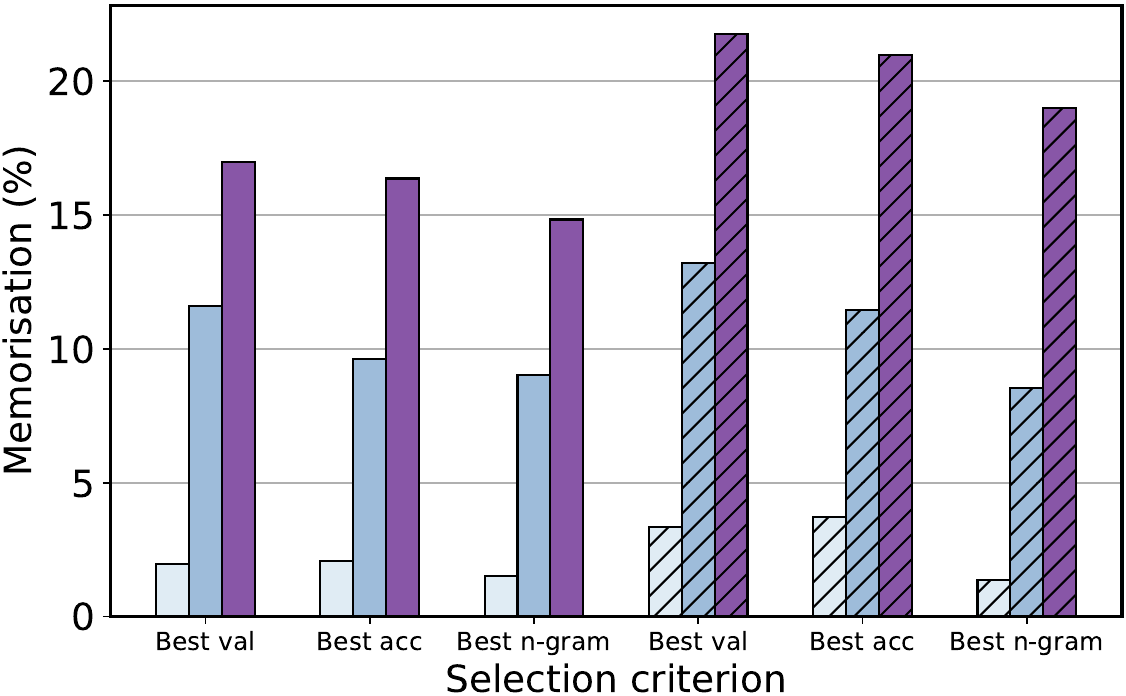}
         \caption{Pythia 12B}
         \label{fig:5c}
     \end{subfigure}
     \hfill
     \begin{subfigure}[t]{0.45\textwidth}
         \centering
         \includegraphics[width=\textwidth]{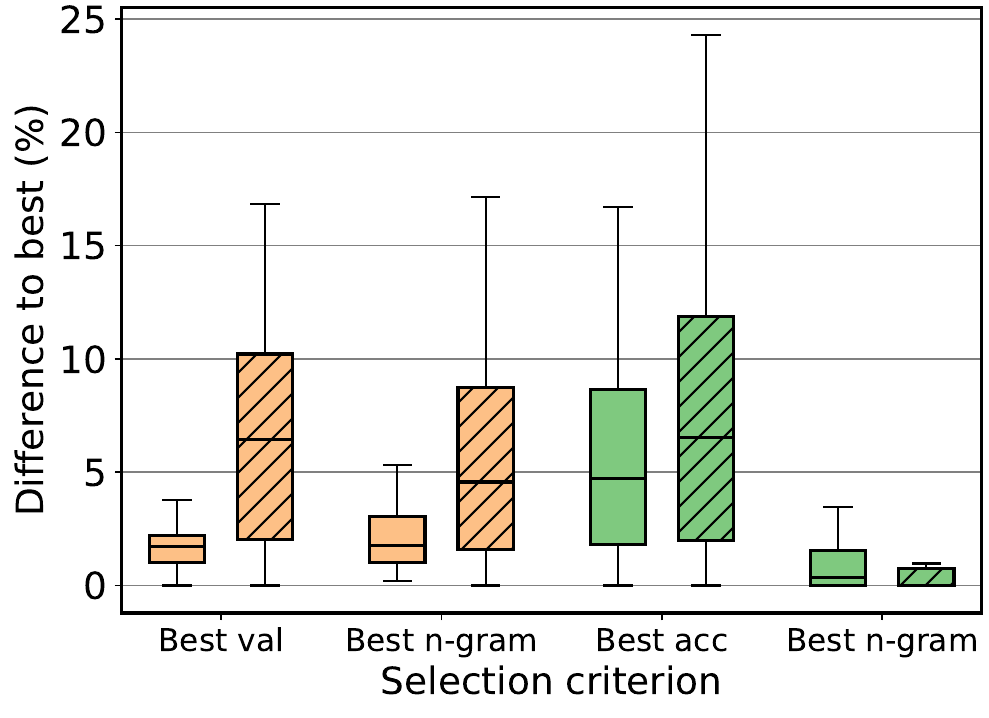}
          \caption{Llama3 8B}
          \label{fig:5d}
     \end{subfigure} \\
    \caption{Memorisation and performance comparison for domain adaptation and instruction tuning across different early stopping selection criteria. (a) and (c) show the verbatim memorisation percentage for different values of extraction prompt prefix length \(k\in\{12,16,20\}\) using three early stopping selection criteria: validation perplexity (Best val), evaluation performance (Best acc), and $n$-gram memorisation (Best $n$-gram) for domain adaptation (solid) and instruction tuning (hatched). (b) and (d) present the difference to the best task evaluation performance (orange) and validation perplexity (green), across the same selection criteria and fine-tuning approaches.}
    \label{fig:5}
\end{figure*}

\subsection{N-gram Memorisation Predicts Verbatim Memorisation}

Driven by the observation shown in Fig. \ref{fig:1} that high-rate memorisation occurs in the early epochs preceding optimal stopping criteria for both validation perplexity and task evaluation performance, we investigate $n$-gram memorisation values as a proxy for fine-grained memorisation. To correctly identify early warning signs of samples at high risk of verbatim memorisation, we evaluate $n$-gram memorisation after each fine-tuning epoch. Fig. \ref{fig:2} shows our results for this evaluation on each of the dataset categories outlined in Table \ref{tab:1}, with \textit{domain} indicating the domain-adaptation fine-tuning. For all samples which are identified as memorised during 8 fine-tuning epochs, we track their associated $n$-gram memorisation score on the epochs preceding the transition to verbatim memorisation. This allows us to understand if the partial memorisation score is higher in the epoch preceding a transition to verbatim memorisation, relative to non-memorised phrases. For this, we plot the average $n$-gram memorisation for non-memorised phrases throughout fine-tuning as a baseline. 

For each of the dataset categories visualised, we observe a clear distinction in partial memorisation between the memorised and non-memorised samples, with the majority of epochs scoring markedly higher than the baseline for non-memorised samples. We see the degree to which this is higher varies significantly between domains, with largest discrepancy observed in Instruction following and Summarisation. The News domains used in the summarisation tend to include high-frequency stock phrases, as such, these datasets are known to encourage extractive copying \citep{tejaswin-etal-2021-well}, which our results agree with.
Most notably, we find that for all datasets, the domain adaptation version sees a significant increase in partial memorisation over the baseline, whereas the baseline scores do not change significantly. Interestingly, there is a large increase in partial memorisation scores of samples which are memorised in the early epochs when performing domain adaptation. 

We perform the same evaluation but compare model size and architecture, shown in Fig. \ref{fig:3}. We identify the expected trend that larger model sizes correlate to higher memorisation capacity, which is reflected in the partial memorisation score increase across the Pythia models. The partial memorisation score gap between memorised and non-memorised samples increases significantly with increasing model size, showing a strong indicator that this metric serves as a scalable precursor to verbatim memorisation.  An unexpected result is that for the smaller 1.4B model, partial memorisation decreases for samples memorised in the latter epochs of fine-tuning; a trend which does not follow for the larger model sizes. Comparing different architectures, we find similar gaps to baseline and the same trend of increasing partial memorisation gap to baseline over fine-tuning epochs.

Figure \ref{fig:4} repeats the analysis for Llama3 8B when only the top 
$n$ transformer layers are updated. The non‑memorised baseline is unaffected, but unfreezing more layers suppresses partial memorisation in the first few epochs and heightens it in later epochs. This is consistent with a capacity‑bottleneck view in which extra trainable layers delay, yet ultimately amplify, overfitting during fine-tuning.

\begin{table*}[t]
  \centering
  \vspace{3pt}
  \resizebox{0.95\textwidth}{!}{
  \begin{tabular}{l l c c c c c c | c c}
    \toprule
    & & \multicolumn{2}{c}{\textbf{QA}} & \multicolumn{2}{c}{\textbf{Summarisation}} & \multicolumn{2}{c}{\textbf{Instruction}} & \multicolumn{2}{|c}{\textbf{Average}} \\
    \cmidrule(lr){3-4}\cmidrule(lr){5-6}\cmidrule(lr){7-8}\cmidrule(lr){9-10}
    \multicolumn{1}{l}{Model + Strategy} && 
      \multicolumn{1}{c}{Mem $\downarrow$} & \multicolumn{1}{c}{Eval $\downarrow$} & 
      \multicolumn{1}{c}{Mem $\downarrow$} & \multicolumn{1}{c}{Eval $\downarrow$} & 
      \multicolumn{1}{c}{Mem $\downarrow$} & \multicolumn{1}{c}{Eval $\downarrow$} &
      \multicolumn{1}{|c}{Mem $\downarrow$} & \multicolumn{1}{|c}{Eval $\downarrow$}\\
    \midrule
    \textbf{Pythia 2.8B}      && 9.71  & - & 12.59 & - & 14.30 & - & 12.20 & - \\
     \quad + Best $n$-gram      && 4.36  & 7.51 & 5.63  & 6.53 & 6.43  & 8.09 & 5.47 & 7.38 \\
     \quad + $n$-gram reg       && \textbf{2.90} & 5.08 & \textbf{3.75} & \textbf{4.25} & \textbf{4.29} & 6.54 & \textbf{3.65} & 5.29 \\
     \quad + Goldfish reg     && 3.37  & \textbf{5.04} & 4.38  & 4.31 & 5.01  & \textbf{6.07} & 4.25 & \textbf{5.14} \\
    \midrule
    \textbf{Pythia 6.9B}      && 12.80 & - & 16.50 & - & 18.70 & - & 16.00 & - \\
     \quad + Best $n$-gram      && 5.76  & 7.54 & 7.42  & 6.21 & 8.42  & 8.30 & 7.20 & 7.35 \\
     \quad + $n$-gram reg       && \textbf{3.84} & 5.15 & \textbf{4.95} & \textbf{4.54} & 5.61  & \textbf{6.30} & \textbf{4.80} & \textbf{5.33} \\
     \quad + Goldfish reg     && 4.48  & \textbf{5.07} & 5.77  & 4.83 & \textbf{5.55} & 6.59 & 5.27 & 5.50 \\
    \midrule
    \textbf{Mistral 7B}       && 13.65 & - & 17.50 & - & 19.88 & - & 17.01 & - \\
     \quad + Best $n$-gram      && 6.12  & 7.55 & 7.88  & 6.00 & 8.91  & 8.89 & 7.64 & 7.48 \\
     \quad + $n$-gram reg       && 4.18  & 5.53 & 5.25  & \textbf{4.40} & 5.34  & \textbf{6.08} & 4.92 & \textbf{5.34} \\
     \quad + Goldfish reg     && \textbf{4.01} & \textbf{5.40} & \textbf{5.12} & 4.42 & \textbf{4.97} & 6.21 & \textbf{4.70} & \textbf{5.34} \\
    \midrule
    \textbf{Llama3 8B}        && 14.40 &- & 18.50 & - & 20.94 & - & 17.95 & - \\
     \quad + Best $n$-gram      && 6.48  & 9.21 & 8.33  & 6.56 & 9.41  & 10.31 & 8.07 & 8.69 \\
     \quad + $n$-gram reg       && \textbf{4.32} & \textbf{4.38} & \textbf{5.55} & \textbf{3.81} & \textbf{6.27} & \textbf{5.32} & \textbf{5.38} & \textbf{4.50} \\
     \quad + Goldfish reg     && 5.04  & 5.02 & 6.47  & 4.33 & 7.32  & 6.99 & 6.28 & 5.45 \\
    \midrule
    \textbf{Pythia 12B}       && 17.66 & - & 22.50 & - & 25.30 & - & 21.82 & - \\
     \quad + Best $n$-gram      && 7.92  & 9.20 & 10.12 & 6.41 & 11.39 & 8.30 & 9.81 & 7.97 \\
     \quad + $n$-gram reg       && \textbf{5.28} & 3.98 & \textbf{6.75} & \textbf{4.02} & \textbf{7.59} & \textbf{4.91} & \textbf{6.54} & \textbf{4.30} \\
     \quad + Goldfish reg     && 6.10  & \textbf{3.90} & 7.57  & 4.36 & 8.86  & 5.00 & 7.51 & 4.42 \\
    \midrule
    \textbf{Llama3 70B}       && 20.80 & - & 26.50 & - & 29.70 & - & 25.67 & - \\
     \quad + Best $n$-gram      && 9.36  & 9.39 & 11.93 & 5.96 & 13.37 & 8.45 & 11.55 & 7.93 \\
     \quad + $n$-gram reg       && \textbf{6.24} & 5.54 & \textbf{7.05} & \textbf{3.91} & \textbf{8.91} & \textbf{5.44} & \textbf{7.40} & \textbf{4.96} \\
     \quad + Goldfish reg     && 7.18  & \textbf{5.50} & 7.27  & 4.01 & 10.40 & 6.11 & 8.28 & 5.21 \\
    \bottomrule
  \end{tabular}}
  \caption{Main memorisation mitigation results across model scales and mitigation strategies. For each result we report the memorisation (\textit{Mem}, lower is better), and Evaluation difference (\textit{Eval}, lower is better) to the best performance achieved for the naive unmitigated strategy (top row of each model group). Bold values indicate the best (lowest) score within each model group (base row excluded). Memorisation scores are taken as the average of all prefix lengths \(k\in\{12,16,20\}\) extractions. Results are averages over 10 randomly initialised fine-tuning runs.}
  \label{tab:2}
\end{table*}

\subsection{Selection Criteria as Mitigation}
\label{sec:4.2}
Following our findings that high-rate memorisation occurs before optimal validation perplexity or task evaluation performance, and that partial memorisation serves as a potential precursor to memorisation, we now investigate the efficacy of utilising this as an early stopping criterion. Without resorting to regularisation or unlearning strategies, we explore using $n$-gram memorisation as a threshold for early stopping. To adapt $n$-gram memorisation as an early stopping criterion, we test different threshold values for which to stop fine-tuning if exceeded. We find that an average partial memorisation threshold score of 20 on the fine-tuning set yields good results. We compare this to the naive selection criterion of validation perplexity and task evaluation for domain adaptation and instruction tuning, respectively, although we experiment with applying validation perplexity and best accuracy to both. 

Results for these experiments are shown in Fig. \ref{fig:5}, highlighting the trade-offs between different early stopping criteria and their impact on both memorisation and model performance. Using evaluation performance/accuracy as the selection criterion consistently reduces memorisation rates in both domain adaptation and instruction tuning scenarios (Fig. \ref{fig:5a} and Fig. \ref{fig:5c}). This could be due to task evaluation performance correlating more highly to the latent capabilities of the pretrained model, rather than validation perplexity on a single domain, and therefore is optimised at lower memorisation. However, this comes at the cost of a significant decrease in validation perplexity, as indicated by the high variance and larger differences to the best perplexity scores shown in Fig. \ref{fig:5b} and Fig. \ref{fig:5d}. Conversely, when validation perplexity is used as the selection criterion, the models tend to show the opposite behaviour through achieving better perplexity scores, but with substantially higher memorisation rates, particularly for instruction-tuned models which consistently exhibit the highest memorisation levels compared to domain adaptation results.

Interestingly, the $n$-gram selection criterion strikes a balance, reducing memorisation without the steep performance trade-offs observed in the other criteria. It provides a more favourable balance by keeping memorisation lower and maintaining better accuracy and perplexity than either of the naive criteria (evaluation accuracy or validation perplexity), as seen by the smaller performance differences at consistently lower memorisation percentages. In summary, instruction tuning appears more prone to memorisation, particularly under validation-based selection, whereas domain adaptation is relatively less affected by these selection criteria, and $n$-gram thresholding as a stopping criterion is a simple and effective memorisation mitigation strategy.

\subsection{Comparing Mitigation Strategies}
We test if our $n$-gram approach can be baked into a loss regularisation function by adapting the typical causal LLM loss to include a term to penalise high-confidence $n$-grams exceeding a tunable confidence threshold, above that of the pretrained model. Intuitively, this penalty is designed to discourage the model from assigning excessively high probabilities to these $n$-grams as a proxy measure for $n$-gram memorisation. The key limitations of this strategy are in requiring the original model to run inference alongside fine-tuning to acquire the baseline confidence values, and keeping $n$-gram sizes within practical bounds to not become computationally intensive. Further details of this approach can be found in Appendix \ref{section:a2}.

We compare our $n$-gram regularisation to the \textit{Goldfish loss} regularisation technique \citep{hans2024like}, incorporating random sampling of dropped tokens from the loss calculation for a given training sample. We test across all models to evaluate transferability and scalability of approach.

We present our results in Table \ref{tab:2}, grouped by model size and dataset category, including comparisons to naive baseline results for both domain adaptation and instruction tuning (top row of each model group). We include the stopping criterion \emph{Best $n$-gram} as a simple non-regularisation approach based on the promising findings in Section \ref{sec:4.2}. We consider memorisation (Mem \%) and evaluation performance (Eval \%), where Eval is taken as the difference to the best achieved performance - essentially measuring the performance trade-off of the memorisation mitigation technique. We group our results by model size, and report the best (bold) within each group. 

\subsubsection{Impact of model size}
These results highlight key trends across different model scales and mitigation strategies. Generally, memorisation increases with model size, as observed with the unmitigated baseline for Pythia 2.8B of 12.2\% rising to 21.8\% for Pythia 12B and 25.7\% for Llama3 70B. Importantly, the mitigation strategies show consistent reductions in memorisation across all models. For example, $n$-gram regularisation reduces memorisation from 12.2\% to 3.6\% in Pythia 2.8B, and from 21.8\% to 6.5\% in Pythia 12B. We see similar reductions in the Llama3 and Mistral models. Goldfish regularisation is also effective, though its impact is more pronounced on the Mistral 7B model, whereas our $n$-gram reg outperforms this on all other models. Across the board, larger models present greater challenges in balancing memorisation, validation perplexity, and accuracy. The results suggest that as model size increases, the trade-offs become more pronounced.

\subsubsection{Impact of mitigation strategy}
Averaged over all models, \textit{n‑gram regularisation} delivers the best trade‑off, lowering memorisation to 5.45\% with a performance evaluation gap of 4.95\%; this is a \(\approx\)40\% relative reduction in memorisation and a \(\approx\)35\% smaller performance hit compared with the simple \textit{Best \(n\)-gram} early‑stopping rule (8.29\%, 7.80\%).  
\textit{Goldfish} is a close second (6.05\%, 5.18\%), performing best on Mistral 7B. While the early‑stopping heuristic of Best $n$-gram consistently sees higher memorisation and worse evaluation performance, it still significantly reduces memorisation from the naive baseline - highlighting the importance of a simple non-regularisation approach.

\begin{figure}[t!]
    \centering
    \includegraphics[width=0.499\textwidth]{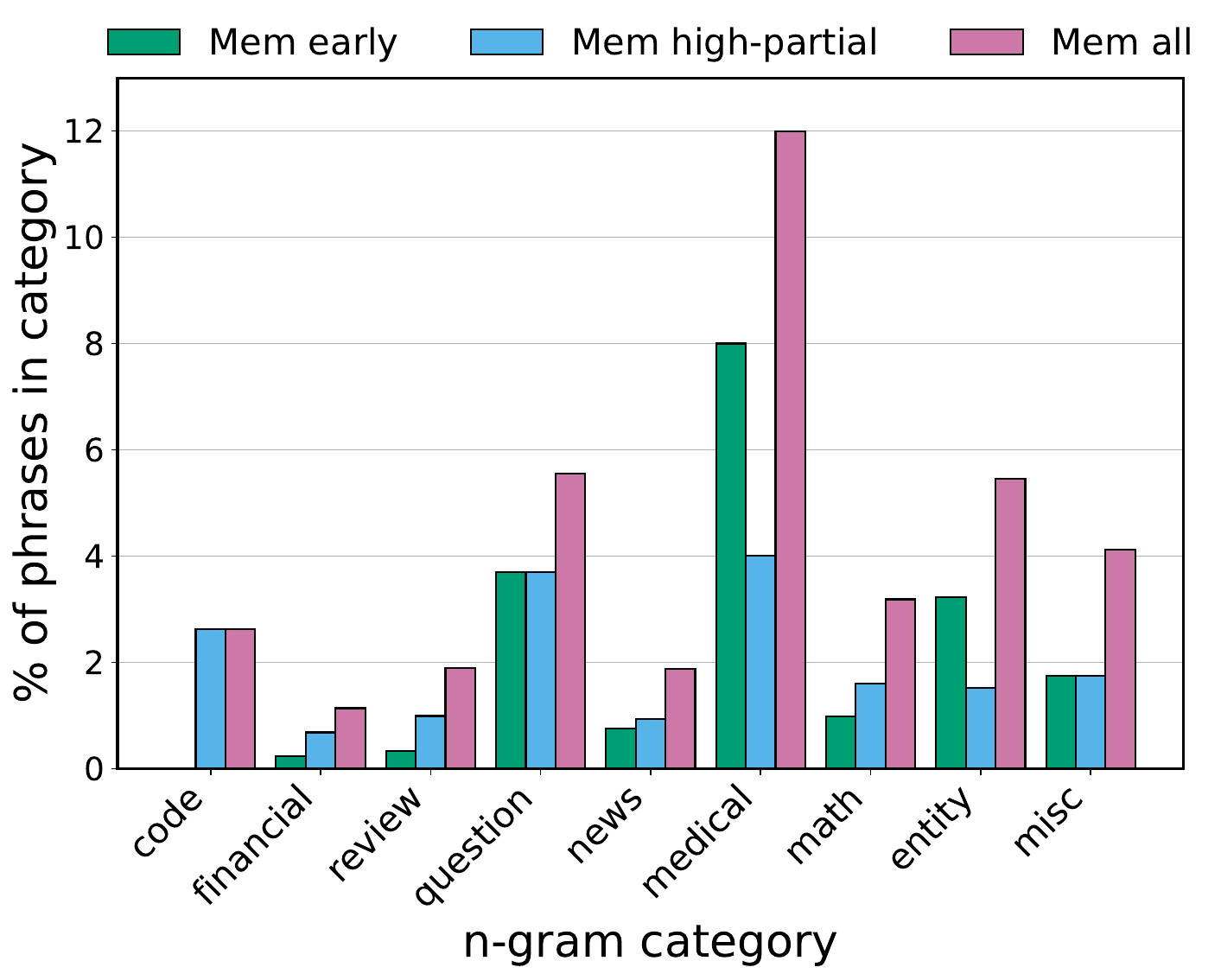}
    \caption{Distribution of memorised $n$-grams across coarse semantic categories.
Bars show the percentage of phrases \emph{within} each category that are: memorised within the first two fine‑tuning epochs and remained memorised (`Mem early', green), memorised after exhibiting higher than baseline partial memorisation in a preceding epoch (`Mem high-partial', blue), and all verbatim memorised phrases (`Mem all', violet).}
    \label{fig:n-gram-analysis}
\end{figure}

\subsubsection{Categorical Analysis}
We construct coarse semantic categories of $n$‑grams over all datasets, randomly sample 500 unique $n$‑grams \emph{per} category, and plot their memorisation outcomes in Figure~\ref{fig:n-gram-analysis}. Memorisation differs markedly between categories, with \emph{medical}, \emph{question}, and \emph{entity} forming the highest‑risk groups, while \emph{financial}, \emph{news}, and \emph{review} remain lowest. Both \emph{medical} and \emph{question} show high rates of early memorised samples that persist, and most categories include many phrases that enter a high‑partial state before verbatim copying, with \emph{code} being the clearest example. Categories whose text is highly templated and repetitive (e.g., medical, questions, and named‑entity lists) likely present many identical $n$‑gram patterns for the model to latch onto, whereas free‑form prose like news or reviews offers far fewer exact repeats. Qualitative examples of memorised phrases are reported in Appendix \ref{section:a3}.

\section{Limitations} Our study provides insights into memorisation during domain adaptation and instruction tuning of causal LLMs, but has limitations. We focused on greedy decoding, while real-world applications often use more complex methods like beam search, which likely influence memorisation differently - future research should explore memorisation under various decoding strategies.

We used validation perplexity and evaluation performance as metrics, but their trade-offs with memorisation aren't necessarily equivalent. Investigating alternative metrics could offer a more nuanced understanding of these relationships. Our experiments were limited to a single high-parameter model (Llama3 70B) due to computational budget limitations - ideally we would evaluate these finding are a larger pool of models and sizes, as well as different fine-tuning protocols.

\section{Conclusion and Future Work}

This study explores memorisation dynamics during both domain adaptation and instruction tuning across eight open‑weight LLMs (1.4B–70B parameters). We show that a simple $n$-gram partial memorisation score indicates at‑risk samples. The gap between memorised and non‑memorised items is widest in domain adaptation and summarisation datasets, reflecting repetition and lack of diversity seen with instruction-tuning, whereas classification and QA tasks exhibit a smaller, but still measurable, rise. We also show that our partial memorisation metric scales very well with increasing model size, where memorisation is more pronounced. Building on these observations, we explore memorisation mitigation strategies. A threshold‑based early stopping with the \(n\)-gram score halves memorisation relative to the baseline at low performance cost, but an explicit \(n\)-gram penalty in the loss is more effective, averaging 5.45\% memorisation and a 4.95\% performance gap: roughly a 40\% reduction in memorisation. We show this scales from small models to 70B‑parameter models and generalises across datasets and tasks.

Future work will extend this analysis in two directions.  First, alternative decoding strategies such as beam search may surface different leakage patterns and should be audited with the same metrics.  Secondly, we will test whether the \(n\)-gram regulariser curbs memorisation in code generation, mathematical reasoning and multimodal tasks. 

\section*{Acknowledgments}
This work was funded by the European Regional Development Fund, and Wordnerds.

\bibliography{mybib}
\bibliographystyle{acl_natbib}
\appendix
\section{Datasets}
\label{section:a1}

The following datasets are used and evaluated according to their respective benchmarks, found in \citet{taori2023alpaca, wei2023flan, wang-etal-2018-glue}.
\begin{itemize}
    \item \textbf{SST‑5}.  
          Movie‑review sentences annotated with five sentiment levels.  
          \textit{Template:} \texttt{Sentence: <s> - What is the sentiment?}.  
          \textbf{Metric:} accuracy.

    \item \textbf{QQP}.  
          Pairs of Quora questions labelled as duplicates or not.  
          \textit{Template:} \texttt{Q1: <q1>\textbackslash nQ2: <q2> - Duplicate? Yes/No}. 
          \textbf{Metric:} accuracy and F1.

    \item \textbf{RTE}.  
          Premise–hypothesis pairs drawn from news and Wikipedia, framed as binary entailment.  
          \textit{Template:} \texttt{Premise: <p> Hypothesis: <h> - Entailed? Yes/No}.  
          \textbf{Metric:} accuracy.

    \item \textbf{WANLI}.  
          Large‑scale adversarial Natural Language Inference corpus generated via human–AI collaboration.  
          \textit{Template:} \texttt{Premise: , Hypothesis:, Label: entail / neutral / contradict}.  
          \textbf{Metric:} accuracy.

    \item \textbf{SQuAD v2}.  
          Wikipedia paragraphs paired with questions, mixing answerable and unanswerable cases.  
          \textit{Template:} \texttt{Context: <para> Question: <q> Answer:}.  
          \textbf{Metric:} exact match (EM) and F1.

    \item \textbf{HellaSwag}.  
          Multiple‑choice commonsense completion task built from WikiHow and activity narratives.  
          \textit{Template:} \texttt{Story: <ctx> Which ending (A–D) is most plausible?}.  
          \textbf{Metric:} multiple‑choice accuracy.

    \item \textbf{PubMedQA‑L}.  
          Biomedical abstracts with yes/no/maybe answers to research questions.  
          \textit{Template:} \texttt{Abstract: <abs> Question: <q> Answer (yes/no/maybe):}. 
          \textbf{Metric:} accuracy.

    \item \textbf{XSum}.  
          BBC news articles paired with single‑sentence abstractive summaries.  
          \textit{Template:} \texttt{Article: <doc> nWrite a one‑sentence summary:}.  
          \textbf{Metric:} ROUGE‑1/2/L.

    \item \textbf{CNN/DailyMail}.  
          Long‑form news articles with multi‑sentence “highlights”.  
          \textit{Template:} \texttt{Article: <doc> Summarise concisely:}.  
          \textbf{Metric:} ROUGE‑1/2/L.

    \item \textbf{Alpaca‑52k}.  
          GPT‑3.5‑generated instruction–response pairs covering diverse tasks.  
          \textit{Template:} \texttt{Instruction: , Input: , Response:}.  
          \textbf{Metric:} GPT‑4 preference win‑rate.

    \item \textbf{FLANv2}.  
          Composite collection of \(\sim\)1.8k tasks (12M examples) in instruction format.  
          \textit{Template:} \texttt{Instruction: \{task\}Input: \{x\} Answer:}.  
          \textbf{Metric:} task‑specific (Accuracy, F1, ROUGE, etc.).
\end{itemize}

\section{N-gram Regularisation Loss}
\label{section:a2}
To incorporate $n$-gram regularisation into the standard causal language modelling loss function, we modify the loss function to include a penalty term that discourages the model from assigning excessively high confidence to certain $n$-grams compared to the pre-trained model. The modified loss function consists of two main components:\\

\noindent\textbf{Primary Loss Term}:
    \begin{align}
    \mathcal{L}_{\text{LM}} = -\sum_{t=1}^{T} \log p_\theta(x_t \mid x_{<t})\nonumber
    \end{align}

\noindent where \( T \) is the total length of the token sequence, \( x_t \) is the token at position \( t \), \( x_{<t} = (x_1, x_2, \ldots, x_{t-1}) \) represents all previous tokens before position \( t \), \( p_\theta(x_t \mid x_{<t}) \) is the probability of token \( x_t \) given previous tokens under the current model parameters \( \theta \). This is the standard cross-entropy loss used for causal LLM training.\\
    
\noindent\textbf{N-gram Regularisation Term}:
\\
\begin{align}
\mathcal{L}_{\text{reg}} = \lambda \sum_{g \in \mathcal{G}}
\bigl[\max\{0,\;p_\theta(g) - p_{\theta_0}(g) - \tau\}\bigr]^2 \nonumber
\end{align}

\noindent  where \( p_\theta(g) \) is the probability assigned by the fine-tuned model to the $n$-gram \( g \), \( p_{\theta_0}(g) \) is the probability assigned by the pre-trained model to the same $n$-gram, \( \lambda \geq 0 \) is the regularisation strength, and \( \tau \geq 0 \) is the confidence margin. For an $n$-gram $g=(w_1,\dots,w_n)$ we compute \(p_\theta(g)=\prod_{i=1}^{n}p_\theta(w_i\mid w_{<i})\) and analogously for \(p_{\theta_0}(g)\).
The penalty is applied only when the fine‑tuned model’s confidence exceeds the pre‑trained model’s by more than \( \tau \); otherwise the term is zero. Balancing this term prevents the model from over‑memorising while preserving latent pre‑trained performance.

\section{Qualitative Examples}
\label{section:a3}
\noindent Table \ref{tab:qual_mem} contains example model output prefix-suffix pairs evaluated as `memorised', alongside any incorrect continuations relative to ground truth.

\begin{table}[t]
\centering
\footnotesize
\begin{tabular}{@{}p{1.6cm}p{5.3cm}@{}}
\toprule
\textbf{Category} & \textbf{Input prefix, predicted suffix, and ground truth} \\
\midrule

Instruction &
\pref{Sophie sat at her desk, staring blankly at the computer} \good{screen. Her mind was racing as she weighed the options} \bad{before deciding to quit her job.}\\
& [GT: in front of her.] \\[6pt]

Review &
\pref{Feels as if there’s a choke leash around your neck} \good{so director Nick Cassavetes can give it a good, hard yank} \bad{and the audience grows impatient.}\\
& [GT: whenever he wants you to feel something.”] \\[6pt]

Question &
\pref{How can I lose ten pounds in three weeks without} \good{exercising? I currently weigh 185 pounds and have an office job} \bad{that requires a lot of sitting.}\\
& [GT: but I dislike counting calories—what should I do?”] \\[6pt]

News &
\pref{An American woman died aboard the MS Veendam, owned by} \good{cruise operator Holland America Line, after the ship docked in Rio de Janeiro} \bad{, officials later added she was travelling alone.}\\
& [GT: on Tuesday, according to the state-run Brazilian news agency Agencia Brasil.”] \\[6pt]

Medical &
\pref{Oral gentamicin as well as oral and intraperitoneal polymyxin B,} %
\good{which binds endotoxin, did not prevent hepatic injury in rats with self‑filling blind loops.}\\
& [GT: \emph{<sentence ends>}] \\[6pt]

Medical &
\pref{However, oral metronidazole and tetracycline therapy continuously administered beginning 1} %
\good{day after surgery diminished hepatic injury (histology score 3.0 +/- 1.8} %
\bad{and led to unexpected weight gain.}\\
& [GT: day after surgery diminished hepatic injury...] \\[6pt]

Financial &
\pref{Dow, S\&P 500 and Nasdaq futures slipped between 0.7\% and} \good{1\% as the rise in bond yields weighed and Apple again fell 1.1\% in pre-market trading ahead of earnings} \bad{while Tesla rose on delivery optimism.}\\
& [GT: \emph{<sentence ends>}] \\[6pt]

Financial  &
\pref{The 30-year bond US30YT=RR firmed 26/32, taking its yield to} \good{4.17 percent, after hitting another record low of 4.16 percent} \bad{as investors fled into equities.}\\
& [GT: \emph{<sentence ends>}] \\[6pt]

\bottomrule
\end{tabular}
\caption{Example predicted continuations given a 10-token input prefix (\pref{grey}). \good{Green} spans mark $\ge$10-token verbatim copies; \bad{red} tokens show divergence. Square-bracketed lines give the Ground-Truth (GT) continuation for each divergent span.}
\label{tab:qual_mem}
\end{table}

\end{document}